\documentclass[11pt]{article}
\usepackage[letterpaper,margin=1in]{geometry}
\usepackage{times}
\usepackage{natbib}
\setcitestyle{authoryear,round,citesep={;},aysep={,},yysep={;}}
\usepackage{microtype}

\usepackage{amsmath,amsfonts,bm}

\def\eqref#1{equation~\ref{#1}}

\def\1{\bm{1}}

\DeclareMathAlphabet{\mathsfit}{\encodingdefault}{\sfdefault}{m}{sl}
\SetMathAlphabet{\mathsfit}{bold}{\encodingdefault}{\sfdefault}{bx}{n}

\usepackage{booktabs}
\usepackage{threeparttable}
\usepackage{xcolor}
\usepackage{multirow}
\usepackage{array}
\usepackage{graphicx}
\usepackage{amssymb}
\usepackage{enumitem}
\usepackage[font=small,labelfont=bf]{caption}

\usepackage{flafter} 
\makeatletter
\newcommand{\flushfloats}{\par\begingroup\let\@elt\relax
  \edef\@tempa{\@deferlist}\ifx\@tempa\@empty\else\newpage\fi\endgroup}
\makeatother
\graphicspath{{figures/}}

\usepackage[colorlinks=true,linkcolor=blue!60!black,citecolor=blue!60!black,urlcolor=blue!60!black]{hyperref}
\usepackage{url}

\newcommand{\cifontsize}{\fontsize{5pt}{5.5pt}\selectfont}
\newcommand{\tablefont}{\fontsize{8pt}{9.6pt}\selectfont\renewcommand{\cifontsize}{\fontsize{6pt}{6.6pt}\selectfont}}
\newcommand{\tablefontsmall}{\fontsize{7pt}{8.4pt}\selectfont\renewcommand{\cifontsize}{\fontsize{5pt}{5.5pt}\selectfont}}
\newcommand{\cifont}{\cifontsize\color{black!60}}
\newcommand{\twolines}[2]{\begin{tabular}[t]{@{}l@{}}#1\\[-1pt]#2\end{tabular}}
\newcommand{\vci}[2]{\begin{tabular}[t]{@{}c@{}}$#1$\\[-2.4pt]{\cifont$[#2]$}\end{tabular}}
\newcommand{\vcib}[2]{\begin{tabular}[t]{@{}c@{}}$\mathbf{#1}$\\[-2.4pt]{\cifont$[#2]$}\end{tabular}}
\newcommand{\vsd}[2]{\begin{tabular}[t]{@{}c@{}}$#1$\\[-2.4pt]{\cifont sd\,$#2$}\end{tabular}}

\title{Diagnosis by Composition:\\ Learning the Visual Vocabulary of Radiological Diagnosis}

\author{Kaihua Hou\textsuperscript{1}, Wei Liu\textsuperscript{2}, Yi Huang\textsuperscript{2}, Hao Zhang\textsuperscript{3}, Haibo Jin\textsuperscript{2}, Yifei Sun\textsuperscript{4}, Yuan Gao\textsuperscript{2},\\
Zuoou Li\textsuperscript{5}, Ling Zhang\textsuperscript{2}, Ahmed Alaa\textsuperscript{1}, Geoffrey H. Tison\textsuperscript{6}, Xu Han\textsuperscript{7}, Ke Yan\textsuperscript{2}\\[0.8em]
\normalsize
\textsuperscript{1}University of California, Berkeley \quad
\textsuperscript{2}Alibaba Group \quad
\textsuperscript{3}Fudan University \quad
\textsuperscript{4}Zhejiang University\\
\textsuperscript{5}Imperial College London \quad
\textsuperscript{6}University of California, San Francisco \quad
\textsuperscript{7}Zhejiang Lab
}
\date{}

\begin{document}

\maketitle
\begin{abstract}

Mainstream CT diagnostic models are trained end-to-end to predict disease labels. However, a disease label is an abstract clinical conclusion rather than a direct visual entity. Instead, CT scans manifest radiological findings, which are observable lesion attributes such as density, margins, and enhancement patterns. Serving as the basis for both clinical guidelines and radiologists' reasoning, these findings constitute the true visual primitives of diagnosis. In this paper, we propose Diagnosis by Composition (DbC), a framework that shifts the learning paradigm from holistic diseases to compositional findings. DbC grounds each finding as a dense activation map aligned with its text embedding, and executes diagnosis by composing the guideline-specified finding maps via a parameter-free deterministic rule. By entirely eliminating disease labels from the training targets, DbC ensures that every prediction is intrinsically interpretable and auditable against clinical definitions. Extensive experiments on multi-phase liver CT demonstrate that DbC learns a more accurate and much more generalizable representation. Under full supervision, DbC outperforms competing methods and exhibits superior data efficiency across all training scales compared to a matched model trained directly on diseases. It further significantly outperforms competing methods in both detection and localization across unseen diseases, rare cross-institutional diseases, and external cohorts. Controlled ablations confirm that these gains stem from learning findings rather than diseases. Overall, DbC shows that learning what is visible and composing what is concluded yields a diagnostic model that is generalizable across diseases and institutions, adaptable to new diseases without retraining, and interpretable and auditable by design.

\end{abstract}
{\bf Code:} \href{https://anonymous.4open.science/r/DiagnosisByComposition}{https://anonymous.4open.science/r/DiagnosisByComposition}


\section{Introduction}
\label{sec:intro}

A clinical diagnosis is an abstract conclusion deduced from an image, rather than a direct visual entity within it. What a CT scan actually manifests are radiological findings, which are observable lesion attributes such as density, margins, and multi-phase contrast enhancement patterns. Radiologists are trained to recognize these findings and synthesize them into a diagnosis \citep{acgme2019drmilestones, hartung2020report, esr2011reporting, acr2025communication}. Similarly, published clinical guidelines define each disease in the exact same terms by specifying the characteristic findings it exhibits \citep{chernyak2018lirads}. Consequently, radiological findings serve as the fundamental, traceable building blocks of diagnosis, constituting its true visual vocabulary.

Mainstream diagnostic models, however, are predominantly trained on the clinical conclusion itself. Under this paradigm, which we term \emph{disease-first}, a model maps an image directly to a disease label \citep{yasaka2018liver, hamm2019liver}. Because the holistic disease is treated as the unit of learning, the underlying visual evidence is either left for the model to implicitly discover or extracted via post-hoc attribution afterwards \citep{wang2019liver}. Recent vision-language models (VLMs), which align images and text in a shared embedding space \citep{radford2021clip}, allow arbitrary diseases to be queried at test time \citep{hamamci2024ctrate, blankemeier2026merlin, du2024segvol, zhao2024sat, rokuss2025voxtell}. Nevertheless, these models still align visual features with high-level disease names or free-text clinical reports. Because routine reports are inherently selective and omit many observable lesion properties, such models lack dedicated representations grounded in individual findings. As a result, their predictions cannot be traced to explicit visual evidence, and their learned representations remain organized around diagnostic conclusions rather than visual primitives.

\begin{figure}[!htb]
\centering
\includegraphics[width=\textwidth]{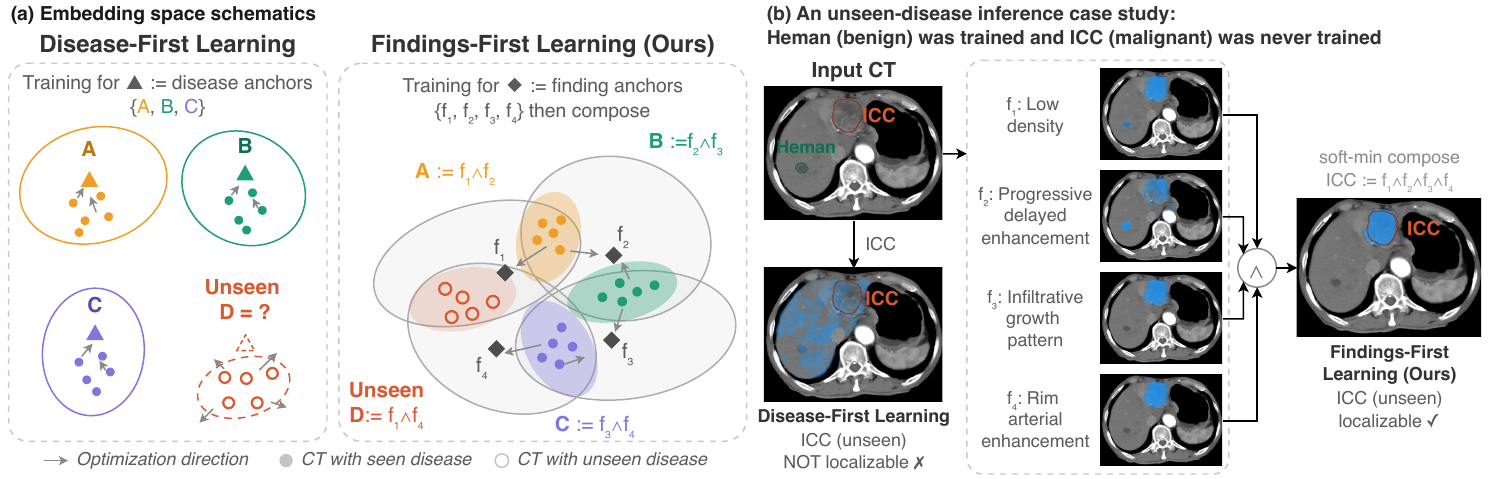}
\caption{\textbf{Disease-first versus findings-first learning.} (a) Disease-first models rely on holistic disease anchors, leaving an unseen disease $D$ without a trained representation, whereas our findings-first approach expresses $D$ as a composition of learned finding anchors. (b) When trained on hemangioma but evaluated on unseen intrahepatic cholangiocarcinoma (ICC), competing methods fail, while DbC accurately localizes the ICC lesion by composing its guideline-defined finding maps.}
\label{fig:fig1}
\end{figure}

This paper asks what a diagnostic model should fundamentally learn to represent, and our answer is the radiological findings themselves. We introduce \textbf{Diagnosis by Composition (DbC)}, a representation learning framework for clinical CT that shifts the unit of learning from diseases to findings (Figure~\ref{fig:fig1}). Grounded in published clinical guidelines and verified by board-certified radiologists, we curate a vocabulary of findings and formalize the compositional definition of each disease over this vocabulary. DbC learns each finding as a dense activation map by aligning local voxel features with the text embedding of the finding name. Crucially, this learning process is supervised both spatially, using lesion masks paired with guideline-deduced findings from disease labels, and semantically, using additional diverse findings extracted from clinical reports. The guideline-deduced findings label every lesion of a disease, and the report findings add what varies from one lesion to the next. By entirely excluding disease labels from the training targets, DbC derives each diagnosis at test time by aggregating the required finding maps via a parameter-free, deterministic rule dictated by clinical guidelines.

Shifting the learning target to findings yields two major advantages. The first is \emph{intrinsic traceability}. Because every diagnosis is a deterministic function of named finding maps, predictions can be audited finding by finding against clinical definitions, and new diseases can be added simply by specifying their required findings. The second is \emph{representation generalizability}. Since findings are shared across pathologies, every labeled lesion informs multiple diagnoses simultaneously. We evaluate DbC across two regimes, namely benchmarking against architecture-matched disease-first counterparts under full supervision, and generalizing to unseen diseases withheld during training via compositional reasoning. The first regime shows whether learning findings costs accuracy on the diseases that a model was trained on. The second is the more demanding test of the representation, because a model must diagnose a disease without a single training example and can reach it only through what it learned on other diseases. Our main contributions are as follows:

\begin{itemize}[leftmargin=*,nosep]
\item We introduce Diagnosis by Composition (DbC), a framework that learns guideline-defined findings as dense, text-aligned maps and executes diagnosis via deterministic clinical rules, ensuring intrinsic traceability by design.
\item We show that learning findings yields more data-efficient and far more generalizable representations. DbC surpasses its disease-first counterpart across all data scales under full supervision, while significantly outperforming competing methods in detecting and localizing unseen, rare, and external-cohort diseases.
\item We verify through controlled ablations that these gains stem directly from the unit of learning. DbC maintains its advantage even when trained solely on weak findings deduced from disease labels, outperforms baseline VLMs prompted with the same findings, and requires no learnable findings-to-disease classifier beyond the deterministic guideline rule.
\end{itemize}


\section{Related work}
\label{sec:related}

\paragraph{Medical vision-language representation learning.}
Contrastive pretraining aligns images and text in a shared embedding space to recognize categories by their names \citep{radford2021clip}. Medical adaptations learn such spaces by pairing biomedical figures \citep{zhang2023biomedclip}, chest radiographs \citep{tiu2022chexzero}, and CT volumes \citep{hamamci2024ctrate, blankemeier2026merlin} with clinical reports, extending alignment to image regions and words \citep{huang2021gloria, wang2022mgca}, clinical entities \citep{wu2023medklip}, and anatomical structures \citep{shui2025fvlm}, while enabling volume-level abnormality scoring without task-specific training examples \citep{lai2025brgsa, park2025radzero3d, lin2025cxrlt}. Language-driven dense prediction further extends this alignment to pixels \citep{li2022language} and 3D volumes \citep{liu2023clipdriven}. Notably, SegVol \citep{du2024segvol}, SAT \citep{zhao2024sat}, and VoxTell \citep{rokuss2025voxtell} segment target regions from text prompts even for unseen diseases. These dense models, alongside our architecture-matched Disease-VL baseline (Section~\ref{sec:setup}), serve as our primary comparators for lesion-level localization. However, across this entire family, diseases remain anchored to holistic disease names or routine reports that omit most observable lesion properties (Section~\ref{sec:supervision}), whereas DbC supervises findings explicitly and densely on lesion masks, with labels from both radiological guidelines and clinical reports. While Malenia \citep{jiang2025malenia} recently incorporates lesion attributes such as shape and density to segment unseen lesion types, all attributes of a candidate region share a single mask and are combined via learned fusion. In contrast, DbC learns a dedicated, independently verifiable spatial map for each finding and derives diagnoses strictly through deterministic guideline rules.

\paragraph{Attribute-based and compositional recognition.}
Recognizing categories through shared semantic attributes is a classic principle in computer vision \citep{lampert2009attribute, xian2019zeroshot}, which has been extended to attribute-object compositions \citep{misra2017composition, nagarajan2018attributes} and dense pixel-level prediction \citep{zhang2024attribute}. DbC builds upon this principle to fundamentally reshape representation learning for clinical diagnosis. Specifically, our visual attributes are the radiological findings defined by clinical guidelines. Each finding is grounded as a dense spatial map, and the attribute-to-class mapping is governed by a deterministic guideline rule evaluated at every voxel. Because findings serve as the sole training targets, this compositional paradigm shapes what the model learns even when all classes are seen, rather than merely dictating how it generalizes to unseen ones.

\paragraph{Concept bottleneck models and description-based inference.}
Concept bottleneck models (CBMs) predict named concepts and learn the mapping from them to each class from examples of that class \citep{koh2020concept}, with the concepts later drawn from a language model or a vision-language space \citep{oikarinen2023labelfree, yuksekgonul2023posthoc, yang2023labo}. DbC shares the named intermediate layer but takes the mapping from findings to a disease from published radiological guidelines, which ties every diagnosis to clinical criteria and defines it for diseases without training examples (Section~\ref{sec:analysis} compares the two). Description-based methods score an image against language descriptions of a class \citep{menon2023visual}. In radiology, Xplainer combines the scores that a pretrained vision-language aligned model gives the observations a radiologist looks for into a zero-shot diagnosis \citep{pellegrini2023xplainer}, CLEAR \citep{han2026clear} scores an image against observations mined from reports, and CDCLIP \citep{li2026cdclip} decomposes a disease into generic visual descriptions that a pretrained CLIP scores. The descriptive terms of these methods are found in reports or scored by a pretrained model, whereas the findings of DbC are training targets that the model learns to localize.


\section{Methods}
\label{sec:method}

\subsection{Findings as the unit of learning}
\label{sec:framework}

Diagnosis by Composition (DbC) mirrors the two-stage reasoning of radiological practice. The model first learns to recognize visual findings within a scan, and subsequently derives each diagnosis by combining the required findings via a deterministic rule translated directly from published clinical guidelines, rather than learning a classifier from data.

Formally, let $x$ be an input CT scan and $\mathcal{F}=\{f_1,\dots,f_M\}$ be a fixed vocabulary of radiological findings, where each $f_m$ represents a distinct, observable visual property. The model predicts a dense \emph{finding map} $a_m(x)\in[0,1]^N$ for each finding, indicating the probability that $f_m$ is present at each of the $N$ spatial tokens of the scan. A token is a small block of voxels (Section~\ref{sec:localizer}). Predicting these finding maps constitutes the sole trainable objective of our framework. Each disease $k$ is formally defined by its \emph{diagnostic composition} $R_k\subseteq\mathcal{F}$, which is the subset of \emph{required findings} specified by its clinical guideline (Section~\ref{sec:vocabulary}). To compute the \emph{diagnosis map} $S_k(x) \in [0,1]^N$, we aggregate the required finding maps using an operator that acts as a continuous relaxation of the logical AND, ensuring that a spatial token scores high for disease $k$ if and only if all of its required findings are simultaneously present. Specifically, we employ the parameter-free \emph{soft minimum} operator, denoted as $\bigwedge$ and applied element-wise across all spatial tokens,
\begin{equation}
S_k(x)=\bigwedge_{m\in R_k} a_m(x)
=\sum_{m\in R_k} a_m(x)\,\frac{\exp\!\big(-\beta\,a_m(x)\big)}{\sum_{j\in R_k}\exp\!\big(-\beta\,a_j(x)\big)},
\label{eq:compose}
\end{equation}
where the temperature parameter $\beta > 0$ controls the strictness of the conjunction (we set $\beta=10$). The soft minimum is a weighted average of the required finding maps in which the smallest value receives the largest weight. A single absent finding therefore keeps the diagnosis map low. Because this aggregation contains no learnable parameters, the resulting diagnosis map preserves the exact semantic meaning of its clinical definition.

At inference time, the scan-level detection score for disease $k$ is obtained via spatial max-pooling over its diagnosis map, $\max_{n} S_{k,n}(x)$, while the connected components of the thresholded map yield the localized lesion predictions. Crucially, unseen diseases without any training examples are diagnosed through the exact same compositional computation, requiring neither architectural modifications nor model retraining. Section~\ref{sec:unseen} explains how each compared model reaches such a disease.

\subsection{Finding vocabulary and diagnostic compositions}
\label{sec:vocabulary}

We instantiate DbC on multi-phase liver CT, where a study may contain any subset of four contrast phases (non-contrast, arterial, portal venous, and delayed), and the model flexibly processes whichever phases are available. Grounded in published clinical guidelines and reference reviews \citep{chernyak2018lirads, galle2018easl, marrero2018aasld} and verified by board-certified abdominal radiologists, we curate a finding vocabulary $\mathcal{F}$ comprising $M=40$ findings organized into 12 semantic categories, including lesion density, margin, and phase-specific enhancement patterns (Table~\ref{tab:compositions}). While $\mathcal{F}$ is constructed to cover the six common liver diseases in our internal cohort, we further evaluate its generalizability on two additional rare diseases in Section~\ref{sec:beyond}.

Under this vocabulary, each modeled disease is formalized by its diagnostic composition $R_k$, requiring two to five characteristic findings that must be simultaneously present (Table~\ref{tab:compositions}). Hepatocellular carcinoma (HCC), for example, typically requires low density on the non-contrast phase, non-rim arterial hyperenhancement, non-peripheral washout and an enhancing capsule. Through the soft minimum operator, $R_k$ serves as the exact clinical rule evaluated at inference for both seen and unseen diseases, while also providing the baseline training supervision for each lesion (Section~\ref{sec:supervision}). Because findings are extensively shared across pathologies, a common primitive such as low density is jointly supervised by the training lesions of every disease that requires it. Figure~\ref{fig:flow} in Appendix~\ref{app:catalog} draws which findings the six diseases share. Detailed definitions, associated contrast phases, and guideline references for all findings and compositions are provided in Appendix~\ref{app:catalog}.

\begin{table}[!htb]
\caption{Diagnostic compositions, each the soft minimum $\wedge$ of Eq.~(\ref{eq:compose}) over the required findings of a disease. NC is the non-contrast phase, and low density on NC is a finding of its own. The two rare diseases were declared at evaluation time from published imaging criteria and never trained on. A circle marks a required finding of exactly one of the six internal diseases, and the last row names the 20 findings that none of the six requires.}
\label{tab:compositions}
\centering
\footnotesize
\begin{tabular}{@{}lp{0.77\textwidth}@{}}
\toprule
Disease & Diagnostic composition (required findings) \\
\midrule
HCC & low density on NC $\wedge$ non-rim arterial hyperenhancement $\wedge$ non-peripheral washout$^{\circ}$ $\wedge$ enhancing capsule$^{\circ}$ \\
ICC & low density $\wedge$ rim arterial enhancement $\wedge$ progressive delayed enhancement $\wedge$ upstream biliary dilatation$^{\circ}$ $\wedge$ infiltrative growth pattern$^{\circ}$ \\
Metastasis & low density $\wedge$ rim arterial enhancement $\wedge$ peripheral washout$^{\circ}$ $\wedge$ ill-defined margin$^{\circ}$ \\
Hemangioma & low density $\wedge$ peripheral nodular discontinuous enhancement$^{\circ}$ $\wedge$ centripetal fill-in$^{\circ}$ $\wedge$ progressive delayed enhancement $\wedge$ well-defined margin \\
FNH & low density on NC $\wedge$ non-rim arterial hyperenhancement $\wedge$ isoenhancing portal phase$^{\circ}$ $\wedge$ well-defined margin $\wedge$ homogeneous enhancement$^{\circ}$ \\
Cyst & water density$^{\circ}$ $\wedge$ no arterial enhancement$^{\circ}$ $\wedge$ well-defined margin $\wedge$ homogeneous density$^{\circ}$ $\wedge$ round or oval shape$^{\circ}$ \\
\midrule
Abscess (rare) & cystic composition $\wedge$ rim arterial enhancement \\
Hepatoblastoma (rare) & low density $\wedge$ non-rim arterial hyperenhancement $\wedge$ non-peripheral washout $\wedge$ heterogeneous density \\
\midrule
Other findings & very high density, iso-density, heterogeneous density, lobulated shape, cystic composition, solid composition, calcified composition, size below 1 cm, size 1 to 2 cm, size 2 to 5 cm, size above 5 cm, central necrosis, intralesional hemorrhage, complete delayed fill-in, delayed central scar enhancement, heterogeneous enhancement, central scar, coarse calcification pattern, targetoid morphology, tumor in vein \\
\bottomrule
\end{tabular}
\end{table}

\subsection{Finding supervisions}
\label{sec:supervision}

Each training lesion is annotated with findings from three complementary sources. First, a lesion is assigned the \emph{required findings} $R_k$ of its disease $k$, regardless of whether they are mentioned in the report. While $R_k$ provides consistent, guideline-grounded labels, relying on it alone assigns identical findings to all instances of a disease, overlooking intra-class variations. Second, to capture such variations, we mine routine clinical reports \citep{irvin2019chexpert, jain2021radgraph, adams2023gpt4} using Qwen3-Max \citep{qwen2025qwen3max} to extract \emph{supportive findings}, which are additional in-vocabulary findings ($\mathcal{F}\setminus R_k$) documented for each lesion. A report may record central necrosis for one lesion of a disease and not for another, for example. Appendix~\ref{app:labels} describes the extraction and its verification. Reports alone are insufficient because radiologists rarely record every observable property. In our data, a report mentions only 1.2 findings per lesion on average, a signal too sparse to learn a complete vocabulary. Thus, guideline-derived required findings ensure comprehensive coverage, while report-extracted supportive findings provide instance-level diversity. Third, each lesion receives one of four \emph{size findings} computed from the physical dimensions of its segmentation mask.

All three finding types supervise the finding maps during training, whereas only $R_k$ forms the diagnostic composition at inference. Finally, lesion masks project these labels onto the spatial grid. For each finding $f_m$, tokens inside any lesion annotated with $f_m$ are positive targets, and all remaining tokens are negatives. A finding label therefore covers its lesion as a whole.

\begin{figure}[!htb]
\centering
\includegraphics[width=0.85\textwidth]{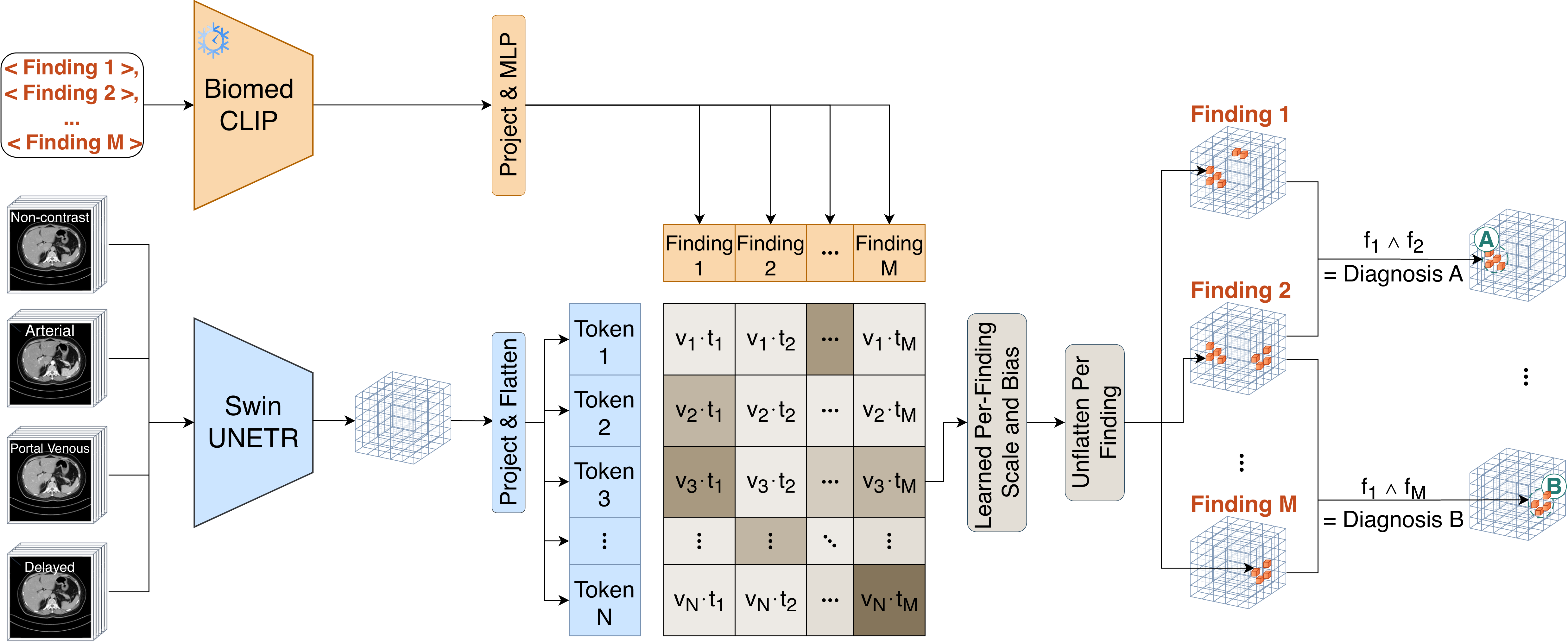}
\caption{The DbC model. Visual embeddings $v_n$ are matched to the text embeddings $t_m$ of the finding names (Eq.~\ref{eq:readout}), and each diagnosis map is the soft minimum of its required finding maps (Eq.~\ref{eq:compose}).}
\label{fig:fig2}
\end{figure}

\subsection{Dense finding localizer}
\label{sec:localizer}

To predict the finding maps (Figure~\ref{fig:fig2}), a four-channel SwinUNETR encoder \citep{hatamizadeh2022swinunetr,he2023swinunetrv2} jointly processes the available contrast phases of an input CT scan $x$, padding any unacquired phase with an all-zero channel. Its dense output feature map is linearly projected and flattened into spatial token embeddings $\{v_n\}_{n=1}^N\subset\mathbb{R}^d$, where each token $n$ corresponds to a local volume of $4\times4\times4$ voxels. In parallel, each finding $f_m\in\mathcal{F}$ is formatted into a fixed text prompt and encoded once by a frozen BiomedCLIP text encoder \citep{zhang2023biomedclip} into a text embedding $t_m$, which a lightweight learnable adapter $\phi$ projects into the $\mathbb{R}^d$ visual token space. The probability $a_{m,n}(x)$ of finding $f_m$ being present at token $n$ is then computed via the scaled cosine similarity between $v_n$ and $\phi(t_m)$ followed by a sigmoid activation $\sigma$,
\begin{equation}
a_{m,n}(x)=\sigma\!\left(s_m\,\frac{v_n^{\top}\phi(t_m)}{\lVert v_n\rVert_2\,\lVert\phi(t_m)\rVert_2}+b_m\right),
\label{eq:readout}
\end{equation}
where $s_m$ and $b_m$ are learnable finding-specific scale and bias parameters. In this manner, each finding map $a_m(x)$ is generated via vision-language matching \citep{radford2021clip} at every spatial token, but supervised with dense voxel-level labels rather than global image-text pairs. Keeping the text encoder frozen preserves pretrained biomedical semantic priors and anchors all findings within a unified language space, allowing diagnostic compositions to flexibly query any finding in $\mathcal{F}$ to define both seen and unseen diseases. All $M=40$ finding maps are jointly optimized end-to-end using a combined Dice and focal loss \citep{milletari2016vnet,lin2017focal}. Appendix~\ref{app:training} gives the prompt, the adapter, the initialization of the readout and the training recipe.


\section{Experiments}
\label{sec:experiments}

We first evaluate DbC under full supervision on seen diseases, followed by evaluations on unseen diseases and cross-institutional cohorts, before conducting controlled ablations to dissect the source of its performance gains and demonstrating how diagnoses are traced to visual findings.

\subsection{Setup}
\label{sec:setup}

\paragraph{Data.}
We curate an internal cohort of 15,609 multi-phase liver CT scans paired with lesion segmentation masks and radiology reports from a single institution, split at the patient level into 12,487 training, 1,559 validation, and 1,563 test scans. Initial lesion masks were delineated by five junior radiologists and subsequently refined and verified by two senior radiologists, with ground-truth disease labels confirmed via electronic medical records. This retrospective study was approved by the institutional review board with a waiver of informed consent. Each scan may contain one or more of six common liver diseases, namely hepatocellular carcinoma (HCC), intrahepatic cholangiocarcinoma (ICC), metastasis (Meta), hemangioma (Heman), focal nodular hyperplasia (FNH), and cyst, while scans without lesions serve as negative controls. Appendix~\ref{app:data} describes the preprocessing and the split.

\paragraph{Regimes.}
A disease is defined as \emph{seen} when included in training, and \emph{unseen} when the model receives neither supervision for it nor exposure to any scan containing it during training. The experiments cover five settings, three on the internal test set and two on cohorts from other institutions.
\begin{itemize}[leftmargin=*,itemsep=2pt,topsep=4pt]
\item \textbf{Seen diseases} (Section~\ref{sec:seen}). All models are trained on the full training split and evaluated on the 1,563 internal test scans.
\item \textbf{Data scaling} (Section~\ref{sec:seen}). DbC and Disease-VL are each trained in three independent runs on nested subsets ranging from 10\% to 100\% of the training data, so that every smaller subset is contained in the larger ones.
\item \textbf{Unseen diseases} (Section~\ref{sec:unseen}). Under the \emph{leave-one-disease-out (LODO)} protocol, for each of the six diseases in turn, we remove all scans containing that disease from the training and validation splits, train a dedicated model on the remaining data, and evaluate it on the held-out scans carrying the withheld disease.
\item \textbf{Rare diseases} (Section~\ref{sec:beyond}). We evaluate the full-training checkpoint on an external cohort comprising two rare liver diseases, abscess and hepatoblastoma, which are completely absent from our internal data.
\item \textbf{External cohort} (Section~\ref{sec:beyond}). On the public MCT-LTDiag cohort \citep{wu2026mctltdiag} containing 482 scans across four of the six common diseases, we evaluate each disease using both the full-training checkpoint (as seen) and the corresponding LODO checkpoint (as unseen).
\end{itemize}

\paragraph{Comparators.}
\label{sec:comparators}
Every comparator is trained on the same CTs, splits and compute as DbC under the same protocol, with its checkpoint selected on the same validation split. We compare with models that return a map, because lesion-level localization tests how precisely a learned representation is grounded in local visual evidence. The comparators form two groups.
\begin{itemize}[leftmargin=*,itemsep=2pt,topsep=4pt]
\item \textbf{Matched ablations} (disease-first, on the encoder of DbC). \textbf{Disease-VL} and \textbf{Disease-V} share the image encoder, input, loss and optimizer of DbC but learn diseases directly, isolating the effect of learning findings-first. \textbf{Disease-VL} aligns the same model with disease names instead of finding names, so it can be prompted with an unseen disease. \textbf{Disease-V} drops the text encoder, a closed-set vision-only control that appears only where every disease is seen. The hyperparameters were searched separately for DbC and for Disease-VL, and the search chose the same values.
\item \textbf{Published baselines} (disease-first vision-language aligned models). SegVol \citep{du2024segvol}, SAT \citep{zhao2024sat} and VoxTell \citep{rokuss2025voxtell} return a region for a text prompt. Each is fine-tuned from its released checkpoint on our data, with a four-phase input layer and its published recipe run for as many steps as DbC (Appendix~\ref{app:prompts}).
\end{itemize}
Disease-VL and the published baselines are prompted with disease names unless stated otherwise.

\paragraph{Metrics.}
We evaluate both lesion-level \emph{localization} (where lesions of a given disease are located) and scan-level \emph{detection} (whether a scan contains the disease). Localization is measured by the lesion-level Dice score, computed between each ground-truth lesion and its best-matching predicted connected component (assigning zero if no prediction overlaps) and averaged across all lesions, ensuring equal weight regardless of lesion size. Detection is measured by scan-level AUROC using the spatial maximum of the diagnosis map, as the scan-level confidence score. For fair comparison, all models are evaluated on a unified spatial grid, and macro-averaged metrics represent the unweighted mean across disease classes. We report 95\% confidence intervals (CIs) estimated via patient-level paired bootstrap with 2,000 replicates, declaring statistical significance when the 95\% CI of the paired difference between two models excludes zero. The tables also report the IoU of the same matched component for localization and the average precision (AP) for detection (Appendix~\ref{app:metrics}). A statement about localization refers to Dice and a statement about detection to AUROC unless a metric is named. Implementation details are given in Appendix~\ref{app:details}, further results in Appendix~\ref{app:results}, and the code is described in Appendix~\ref{app:code}.

\subsection{Seen diseases and data scaling}
\label{sec:seen}

Learning findings does not reduce accuracy on seen diseases, and instead improves it. With all six diseases seen in training, DbC outperforms all five comparators in both localization and detection (Table~\ref{tab:seen}). It is the most accurate of the six models on all four metrics. Because the matched baselines (Disease-VL and Disease-V) share the exact backbone, training data, and recipe of DbC while training directly on disease masks, this improvement does not come from the network architecture, and Section~\ref{sec:analysis} shows that it does not come from the additional report-extracted finding labels either.

\begin{table}[!htb]
\caption{Seen diseases, with every model trained on the full training data (macro means, Appendix~\ref{app:metrics}). The 95\% confidence interval is under each value. Bold marks the best value and $^{\dagger}$ a matched ablation.}
\label{tab:seen}
\centering
\tablefont
\setlength{\tabcolsep}{7pt}
\begin{tabular}[t]{@{}lcccc@{}}
\toprule
& \multicolumn{2}{c}{Localization} & \multicolumn{2}{c}{Detection} \\
\cmidrule(lr){2-3}\cmidrule(l){4-5}
Model & Dice & IoU & AUROC & AP \\
\midrule
SegVol & \vci{.055}{.039,.076} & \vci{.037}{.026,.053} & \vci{.806}{.759,.848} & \vci{.284}{.213,.377} \\
SAT & \vci{.190}{.156,.228} & \vci{.152}{.124,.184} & \vci{.900}{.888,.912} & \vci{.528}{.484,.582} \\
VoxTell & \vci{.096}{.075,.124} & \vci{.069}{.054,.091} & \vci{.810}{.794,.825} & \vci{.230}{.215,.258} \\
Disease-V$^{\dagger}$ & \vci{.288}{.255,.326} & \vci{.233}{.205,.266} & \vci{.915}{.903,.928} & \vci{.628}{.578,.678} \\
Disease-VL$^{\dagger}$ & \vci{.271}{.235,.311} & \vci{.220}{.189,.255} & \vci{.916}{.904,.927} & \vci{.582}{.538,.642} \\
\midrule
DbC (ours) & \vcib{.313}{.281,.349} & \vcib{.253}{.226,.284} & \vcib{.931}{.920,.941} & \vcib{.663}{.615,.708} \\
\bottomrule
\end{tabular}

\end{table}

\begin{figure}[!htb]
\centering
\includegraphics[width=0.66\textwidth]{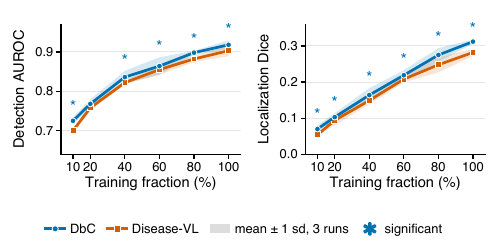}
\caption{Data scaling, DbC against Disease-VL over three runs per training-set size. A star marks a significant advantage of DbC.}
\label{fig:scaling}
\end{figure}

In the data-scaling study across six training-set sizes from 10\% to 100\% (Figure~\ref{fig:scaling}), where each point is the mean of three runs, both models improve with more data and DbC significantly outperforms Disease-VL at almost every scale. Each run draws its own subset and its own initialization, and every run is scored on the same internal test set. The relative gain is largest at the smallest data scale (10\%), supporting the view that shared findings use each labeled lesion more efficiently than isolated disease labels. Appendix~\ref{app:scaling} lists the plotted values and reports the same comparison for the malignant subset.

\flushfloats
\subsection{Generalization to unseen diseases}
\label{sec:unseen}

With accuracy on seen diseases established, we next evaluate generalization to unseen diseases under the LODO protocol, which directly tests the quality of the learned representation \citep{radford2021clip}. DbC and every comparator in Table~\ref{tab:unseen} are vision-language aligned models, and all of them reach an unseen disease through the principle that lets CLIP recognize categories it was never trained on. CLIP embeds images and text in a shared space and recognizes a category by matching an image to the text embedding of its name, so that in a well-aligned space an image of a new category lands near its name although the two were never paired. The models compared here make this match at every location, producing a map rather than a single score, and none is retrained for the unseen disease. What differs between them is the text that the image is matched with, the anchors of Figure~\ref{fig:fig1}a. A disease-first model matches image features with the embedding of the name of the unseen disease, while DbC matches them with the embeddings of its required findings and composes the resulting maps with the soft minimum.

Withholding a disease also removes labels from some of its findings. Notably, for every disease, at least one required finding belongs to none of the other five diagnostic compositions (marked with circles in Table~\ref{tab:compositions}) and thus has few or no labeled examples once that disease is withheld. Consequently, each LODO model must compose the unseen disease partly from shared findings learned on seen diseases, and partly from unseen or rarely labeled findings reached through the shared text space.

\begin{table}[!htb]
\caption{Diagnosis of unseen diseases under leave-one-disease-out, with localization above and detection below. Notation as in Table~\ref{tab:seen}, and $^{*}$ marks a significant advantage over every comparator.}
\label{tab:unseen}
\centering
\tablefontsmall
\setlength{\tabcolsep}{2pt}
\makebox[\textwidth][c]{\begin{tabular}{@{}lcccccccccccc@{}}
\toprule
Model & \multicolumn{2}{c}{HCC} & \multicolumn{2}{c}{ICC} & \multicolumn{2}{c}{Metastasis} & \multicolumn{2}{c}{Hemangioma} & \multicolumn{2}{c}{FNH} & \multicolumn{2}{c}{Cyst} \\
\cmidrule(lr){2-3}\cmidrule(lr){4-5}\cmidrule(lr){6-7}\cmidrule(lr){8-9}\cmidrule(lr){10-11}\cmidrule(l){12-13}
\emph{Localization} & Dice & IoU & Dice & IoU & Dice & IoU & Dice & IoU & Dice & IoU & Dice & IoU \\
\midrule
SegVol & \vci{.035}{.027,.043} & \vci{.020}{.016,.025} & \vci{.065}{.057,.074} & \vci{.037}{.032,.042} & \vci{.021}{.017,.026} & \vci{.012}{.010,.015} & \vci{.051}{.047,.055} & \vci{.029}{.027,.032} & \vci{.048}{.029,.075} & \vci{.028}{.016,.045} & \vci{.001}{.001,.002} & \vci{.001}{.000,.001} \\
SAT & \vci{.215}{.185,.247} & \vci{.161}{.138,.188} & \vci{.206}{.180,.238} & \vci{.147}{.126,.169} & \vci{.147}{.127,.170} & \vci{.110}{.093,.128} & \vci{.241}{.227,.257} & \vci{.176}{.165,.188} & \vci{.150}{.098,.223} & \vci{.106}{.067,.160} & \vci{.032}{.028,.037} & \vci{.024}{.020,.027} \\
VoxTell & \vci{.206}{.175,.240} & \vci{.157}{.133,.182} & \vci{.292}{.258,.329} & \vci{.222}{.195,.251} & \vci{.013}{.008,.019} & \vci{.009}{.005,.013} & \vci{.094}{.087,.102} & \vci{.059}{.053,.064} & \vci{.145}{.100,.207} & \vci{.093}{.062,.135} & \vci{.001}{.000,.002} & \vci{.001}{.000,.001} \\
Disease-VL$^{\dagger}$ & \vci{.249}{.217,.283} & \vci{.195}{.168,.225} & \vci{.347}{.315,.378} & \vci{.263}{.240,.290} & \vci{.098}{.082,.117} & \vci{.074}{.062,.090} & \vci{.133}{.121,.146} & \vci{.099}{.090,.110} & \vci{.035}{.017,.061} & \vci{.022}{.009,.041} & \vci{.015}{.011,.018} & \vci{.010}{.008,.012} \\
\addlinespace[3pt]
DbC (ours) & \vcib{.376^{*}}{.328,.423} & \vcib{.311^{*}}{.271,.353} & \vcib{.394^{*}}{.358,.433} & \vcib{.321^{*}}{.292,.354} & \vcib{.165^{*}}{.143,.192} & \vcib{.131^{*}}{.112,.154} & \vcib{.332^{*}}{.315,.351} & \vcib{.266^{*}}{.252,.282} & \vcib{.265^{*}}{.182,.375} & \vcib{.212^{*}}{.140,.305} & \vcib{.059^{*}}{.053,.065} & \vcib{.042^{*}}{.037,.046} \\
\midrule
\emph{Detection} & AUROC & AP & AUROC & AP & AUROC & AP & AUROC & AP & AUROC & AP & AUROC & AP \\
\midrule
SegVol & \vci{.757}{.735,.779} & \vci{.560}{.522,.600} & \vci{.835}{.816,.853} & \vci{.447}{.406,.491} & \vci{.731}{.706,.756} & \vci{.461}{.421,.504} & \vci{.764}{.746,.783} & \vci{.705}{.677,.733} & \vci{.796}{.756,.832} & \vci{.135}{.103,.177} & \vci{.734}{.716,.752} & \vci{.735}{.710,.761} \\
SAT & \vci{.809}{.790,.827} & \vci{.607}{.566,.648} & \vci{.804}{.782,.824} & \vci{.420}{.381,.463} & \vci{.828}{.809,.846} & \vci{.586}{.541,.632} & \vci{.714}{.695,.733} & \vci{.724}{.699,.749} & \vci{.511}{.457,.564} & \vci{.061}{.045,.091} & \vci{.803}{.786,.819} & \vci{.807}{.785,.827} \\
VoxTell & \vci{.811}{.792,.829} & \vci{.528}{.496,.561} & \vci{.870}{.854,.885} & \vci{.501}{.460,.547} & \vci{.724}{.701,.746} & \vci{.405}{.373,.439} & \vci{.807}{.789,.824} & \vci{.783}{.757,.808} & \vci{.713}{.674,.753} & \vci{.089}{.071,.110} & \vci{.775}{.757,.793} & \vci{.750}{.725,.776} \\
Disease-VL$^{\dagger}$ & \vci{.691}{.655,.723} & \vcib{.624}{.575,.674} & \vci{.828}{.799,.856} & \vci{.711}{.659,.764} & \vci{.765}{.730,.797} & \vci{.617}{.558,.678} & \vci{.776}{.747,.804} & \vcib{.822}{.790,.852} & \vci{.532}{.476,.585} & \vci{.121}{.096,.152} & \vci{.689}{.655,.721} & \vcib{.808}{.775,.839} \\
\addlinespace[3pt]
DbC (ours) & \vcib{.834^{*}}{.817,.851} & \vci{.577}{.542,.614} & \vcib{.932^{*}}{.920,.944} & \vcib{.767^{*}}{.720,.810} & \vcib{.846^{*}}{.828,.862} & \vcib{.624}{.579,.667} & \vcib{.839^{*}}{.823,.855} & \vci{.801}{.777,.825} & \vcib{.801}{.764,.837} & \vcib{.165}{.118,.231} & \vcib{.809}{.793,.825} & \vci{.798}{.774,.821} \\
\bottomrule
\end{tabular}
}
\end{table}

DbC localizes all six unseen diseases significantly better than every comparator, and achieves the highest detection AUROC on all six, with statistically significant gains on four (Table~\ref{tab:unseen}). The localization advantage is significant by Dice and by IoU. The comparison with Disease-VL isolates the effect of learning findings versus learning diseases under the exact same backbone, training data, and recipe. Disease-VL detects FNH with an AUROC of $0.532$ (near chance) and HCC with $0.691$, and localizes cyst with a Dice of only $0.015$. Cyst is the hardest to localize across all models due to its small lesion size (Appendix~\ref{app:size}). These results reflect a direct advantage of the learning target. Unlike a disease name that states a clinical conclusion, findings describe directly visible properties that recur across diseases, allowing visual evidence learned on seen diseases to transfer directly to unseen ones.

\flushfloats
\subsection{Rare diseases and an external cohort}
\label{sec:beyond}

Evaluating on rare cross-institutional diseases places an even stricter test on the finding vocabulary than LODO. Because abscess and hepatoblastoma are absent from all internal splits and the vocabulary $\mathcal{F}$ was constructed solely for the six common diseases (Section~\ref{sec:vocabulary}), this experiment tests whether $\mathcal{F}$ generalizes to pathologies it was never designed to cover. Following the same protocol used for the common diseases, their diagnostic compositions were defined from published imaging criteria independently of the model and its outputs (Table~\ref{tab:compositions}). Each of the two compositions includes a finding that none of the six common diseases requires, namely cystic composition for abscess and heterogeneous density for hepatoblastoma. DbC evaluates the two compositions with the checkpoint trained on the six common diseases, and the comparators are prompted with the names of the two diseases. On both rare diseases, DbC significantly outperforms every comparator in both localization and detection (Table~\ref{tab:rare}), and the advantage is significant on all four metrics.

\begin{table}[!htb]
\caption{Rare diseases from a different institution. Notation as in Table~\ref{tab:unseen}.}
\label{tab:rare}
\centering
\tablefont
\setlength{\tabcolsep}{4pt}
\begin{tabular}{@{}lcccccccc@{}}
\toprule
& \multicolumn{4}{c}{Abscess} & \multicolumn{4}{c}{Hepatoblastoma} \\
\cmidrule(lr){2-5}\cmidrule(l){6-9}
& \multicolumn{2}{c}{Localization} & \multicolumn{2}{c}{Detection} & \multicolumn{2}{c}{Localization} & \multicolumn{2}{c}{Detection} \\
\cmidrule(lr){2-3}\cmidrule(lr){4-5}\cmidrule(lr){6-7}\cmidrule(l){8-9}
Model & Dice & IoU & AUROC & AP & Dice & IoU & AUROC & AP \\
\midrule
SegVol & \vci{.051}{.042,.060} & \vci{.027}{.022,.033} & \vci{.815}{.767,.863} & \vci{.751}{.676,.826} & \vci{.106}{.063,.155} & \vci{.070}{.039,.104} & \vci{.634}{.543,.719} & \vci{.290}{.186,.426} \\
SAT & \vci{.174}{.119,.231} & \vci{.097}{.063,.140} & \vci{.789}{.740,.836} & \vci{.782}{.721,.837} & \vci{.142}{.095,.199} & \vci{.092}{.059,.130} & \vci{.467}{.392,.542} & \vci{.222}{.149,.295} \\
VoxTell & \vci{.094}{.069,.123} & \vci{.066}{.048,.086} & \vci{.781}{.731,.829} & \vci{.799}{.742,.848} & \vci{.118}{.075,.169} & \vci{.071}{.044,.105} & \vci{.662}{.605,.721} & \vci{.306}{.260,.360} \\
Disease-VL$^{\dagger}$ & \vci{.170}{.135,.211} & \vci{.113}{.087,.145} & \vci{.830}{.794,.861} & \vci{.840}{.800,.893} & \vci{.053}{.027,.086} & \vci{.032}{.015,.053} & \vci{.558}{.444,.665} & \vci{.254}{.172,.376} \\
\midrule
DbC (ours) & \vcib{.238^{*}}{.203,.284} & \vcib{.163^{*}}{.136,.196} & \vcib{.890^{*}}{.851,.924} & \vcib{.865^{*}}{.807,.916} & \vcib{.285^{*}}{.222,.355} & \vcib{.185^{*}}{.142,.235} & \vcib{.829^{*}}{.772,.883} & \vcib{.456^{*}}{.207,.595} \\
\bottomrule
\end{tabular}

\end{table}

The second cross-institutional evaluation transfers the common diseases to the public MCT-LTDiag cohort, which holds four of the six and on which no weight or prompt was fit. Each disease is scored in two ways, in zero-shot transfer with the LODO checkpoint that never saw it, and after full training with the checkpoint trained on all six diseases. Here DbC again significantly outperforms all comparators in localization and detection (by Dice, IoU and AUROC), both in zero-shot transfer and after full training (Table~\ref{tab:mct}). Appendix~\ref{app:mct} reports these results for each of the four diseases. Thus, the advantage of learning findings holds both beyond the diseases the vocabulary was built for and across institutions.

\begin{table}[!htb]
\caption{The public MCT-LTDiag cohort (macro mean over its four diseases), zero-shot and after full training. Notation as in Table~\ref{tab:unseen}.}
\label{tab:mct}
\centering
\tablefont
\setlength{\tabcolsep}{4pt}
\begin{tabular}{@{}lcccccccc@{}}
\toprule
& \multicolumn{4}{c}{Zero-shot} & \multicolumn{4}{c}{Full training} \\
\cmidrule(lr){2-5}\cmidrule(l){6-9}
& \multicolumn{2}{c}{Localization} & \multicolumn{2}{c}{Detection} & \multicolumn{2}{c}{Localization} & \multicolumn{2}{c}{Detection} \\
\cmidrule(lr){2-3}\cmidrule(lr){4-5}\cmidrule(lr){6-7}\cmidrule(l){8-9}
Model & Dice & IoU & AUROC & AP & Dice & IoU & AUROC & AP \\
\midrule
SegVol & \vci{.044}{.039,.051} & \vci{.026}{.023,.030} & \vci{.489}{.458,.520} & \vci{.245}{.232,.264} & \vci{.133}{.119,.150} & \vci{.098}{.087,.111} & \vci{.614}{.586,.642} & \vci{.381}{.351,.419} \\
SAT & \vci{.094}{.083,.109} & \vci{.065}{.056,.076} & \vci{.528}{.497,.559} & \vci{.281}{.259,.313} & \vci{.088}{.076,.101} & \vci{.061}{.052,.071} & \vci{.596}{.569,.623} & \vci{.350}{.324,.387} \\
VoxTell & \vci{.064}{.055,.075} & \vci{.042}{.036,.050} & \vci{.466}{.441,.490} & \vci{.238}{.227,.254} & \vci{.134}{.121,.151} & \vci{.094}{.085,.107} & \vci{.601}{.575,.627} & \vci{.370}{.341,.406} \\
Disease-VL$^{\dagger}$ & \vci{.103}{.091,.119} & \vci{.073}{.065,.085} & \vci{.593}{.560,.625} & \vci{.278}{.263,.302} & \vci{.183}{.168,.205} & \vci{.142}{.130,.160} & \vci{.716}{.692,.740} & \vci{.516}{.487,.552} \\
\midrule
DbC (ours) & \vcib{.173^{*}}{.156,.195} & \vcib{.134^{*}}{.120,.152} & \vcib{.639^{*}}{.606,.673} & \vcib{.326}{.306,.358} & \vcib{.286^{*}}{.269,.310} & \vcib{.247^{*}}{.233,.267} & \vcib{.743^{*}}{.718,.766} & \vcib{.571^{*}}{.534,.616} \\
\bottomrule
\end{tabular}

\end{table}

\flushfloats
\subsection{What drives the gain}
\label{sec:analysis}

Three controlled experiments examine the source of DbC's performance advantage (Tables~\ref{tab:analysisseen} and~\ref{tab:analysisunseen}). Their setups are described in Appendices~\ref{app:training} and~\ref{app:prompts}.

\paragraph{Same supervision, learned as findings.}
Full DbC uses report-extracted supportive findings and mask-derived size findings that disease-first models do not receive. To isolate the effect of the learning unit, we train a \emph{required-only} DbC using solely the required findings $R_k$ (the rows ``DbC, required findings only'' of Tables~\ref{tab:analysisseen} and~\ref{tab:analysisunseen}). Because $R_k$ is deterministically mapped from each lesion's disease label, this variant and the matched baselines learn from the exact same masks and disease supervision, differing only in whether the targets are findings or diseases. On seen diseases, required-only DbC performs close to the full model and still outperforms both matched baselines on all four metrics (Table~\ref{tab:seen}). On every unseen disease, it significantly outperforms Disease-VL in both localization and detection ($^{\diamond}$ in Table~\ref{tab:analysisunseen}). Thus, supportive and size findings account for little of the gain over the matched baselines. Instead, their primary role is to expand the learned vocabulary. Each of the two rare-disease compositions relies on a finding supervised exclusively by clinical reports.

\begin{table}[!htb]
\caption{Analysis of the gain on seen diseases (macro means). The disease-first models are prompted with the required findings of each disease instead of its name. Notation as in Table~\ref{tab:seen}.}
\label{tab:analysisseen}
\centering
\tablefont
\setlength{\tabcolsep}{7pt}
\begin{tabular}[t]{@{}lcccc@{}}
\toprule
& \multicolumn{2}{c}{Localization} & \multicolumn{2}{c}{Detection} \\
\cmidrule(lr){2-3}\cmidrule(l){4-5}
Model & Dice & IoU & AUROC & AP \\
\midrule
\multicolumn{5}{@{}l}{\emph{Disease-first, prompted with finding descriptions}} \\
SegVol & \vci{.033}{.022,.048} & \vci{.020}{.013,.031} & \vci{.791}{.774,.808} & \vci{.231}{.211,.264} \\
SAT & \vci{.196}{.163,.234} & \vci{.149}{.123,.180} & \vci{.819}{.803,.833} & \vci{.261}{.225,.303} \\
VoxTell & \vci{.084}{.067,.109} & \vci{.061}{.048,.079} & \vci{.809}{.794,.825} & \vci{.230}{.213,.255} \\
Disease-VL$^{\dagger}$ & \vci{.273}{.240,.313} & \vci{.216}{.190,.248} & \vci{.873}{.861,.886} & \vci{.354}{.332,.396} \\
\midrule
\multicolumn{5}{@{}l}{\emph{Findings-first}} \\
DbC, learned composition & \vci{.301}{.270,.338} & \vci{.243}{.216,.274} & \vci{.930}{.920,.940} & \vci{.635}{.585,.682} \\
DbC, required findings only & \vci{.307}{.271,.347} & \vci{.248}{.218,.283} & \vci{.929}{.916,.940} & \vci{.651}{.599,.703} \\
DbC (ours) & \vcib{.313}{.281,.349} & \vcib{.253}{.226,.284} & \vcib{.931}{.920,.941} & \vcib{.663}{.615,.708} \\
\bottomrule
\end{tabular}

\end{table}

\paragraph{Findings in the test prompt.}
A disease-first model can also be provided with findings at inference. Keeping model weights fixed, we prompt Disease-VL and the published baselines with text descriptions listing the required findings of each disease, matching the exact findings composed by DbC. For HCC, for example, the published baselines receive the description ``liver lesion with low density on the non-contrast phase, non-rim arterial hyperenhancement, non-peripheral washout and enhancing capsule''. Every model prompted this way still falls short of DbC on seen diseases and on all unseen diseases, where DbC localizes every unseen disease significantly better and achieves higher detection AUROC on all of them. Supplying findings at test time therefore does not close the gap, confirming that the advantage stems from learning representations directly on findings during training.

\begin{table}[!htb]
\caption{Analysis of the gain on unseen diseases, with localization above and detection below. SegVol, SAT, VoxTell and Disease-VL are prompted with the required findings of each disease instead of its name. Notation as in Table~\ref{tab:seen}. $^{*}$ marks a significant advantage of DbC over these four models, and $^{\diamond}$ a significant advantage of the row DbC, required findings only, over Disease-VL prompted with disease names (Table~\ref{tab:unseen}). The learned composition has no output for an unseen disease.}
\label{tab:analysisunseen}
\centering
\tablefontsmall
\setlength{\tabcolsep}{2pt}
\makebox[\textwidth][c]{\begin{tabular}{@{}lcccccccccccc@{}}
\toprule
Model & \multicolumn{2}{c}{HCC} & \multicolumn{2}{c}{ICC} & \multicolumn{2}{c}{Metastasis} & \multicolumn{2}{c}{Hemangioma} & \multicolumn{2}{c}{FNH} & \multicolumn{2}{c}{Cyst} \\
\cmidrule(lr){2-3}\cmidrule(lr){4-5}\cmidrule(lr){6-7}\cmidrule(lr){8-9}\cmidrule(lr){10-11}\cmidrule(l){12-13}
\emph{Localization} & Dice & IoU & Dice & IoU & Dice & IoU & Dice & IoU & Dice & IoU & Dice & IoU \\
\midrule
SegVol & \vci{.048}{.041,.058} & \vci{.028}{.023,.034} & \vci{.070}{.061,.079} & \vci{.039}{.034,.045} & \vci{.013}{.009,.017} & \vci{.007}{.005,.010} & \vci{.048}{.044,.053} & \vci{.028}{.025,.030} & \vci{.048}{.028,.075} & \vci{.028}{.016,.044} & \vci{.004}{.003,.006} & \vci{.002}{.002,.003} \\
SAT & \vci{.259}{.222,.296} & \vci{.203}{.173,.234} & \vci{.332}{.297,.370} & \vci{.252}{.224,.282} & \vci{.151}{.132,.174} & \vci{.112}{.097,.130} & \vci{.260}{.244,.276} & \vci{.203}{.191,.217} & \vci{.151}{.065,.237} & \vci{.103}{.057,.170} & \vci{.030}{.025,.035} & \vci{.021}{.018,.025} \\
VoxTell & \vci{.159}{.134,.188} & \vci{.122}{.102,.144} & \vci{.273}{.240,.308} & \vci{.206}{.181,.234} & \vci{.010}{.006,.015} & \vci{.006}{.004,.009} & \vci{.158}{.145,.170} & \vci{.116}{.107,.126} & \vci{.210}{.135,.306} & \vci{.139}{.092,.203} & \vci{.000}{.000,.000} & \vci{.000}{.000,.000} \\
Disease-VL$^{\dagger}$ & \vci{.019}{.013,.028} & \vci{.013}{.008,.019} & \vci{.315}{.280,.353} & \vci{.244}{.216,.274} & \vci{.090}{.076,.108} & \vci{.069}{.057,.083} & \vci{.075}{.067,.084} & \vci{.049}{.044,.055} & \vci{.168}{.107,.251} & \vci{.129}{.081,.196} & \vci{.005}{.004,.007} & \vci{.003}{.002,.005} \\
\addlinespace[3pt]
\twolines{DbC, required}{findings only} & \vci{.357^{\diamond}}{.305,.408} & \vci{.294^{\diamond}}{.251,.339} & \vci{.387^{\diamond}}{.347,.429} & \vci{.312}{.281,.348} & \vci{.155^{\diamond}}{.131,.184} & \vci{.122^{\diamond}}{.102,.147} & \vci{.302^{\diamond}}{.284,.323} & \vci{.241^{\diamond}}{.226,.258} & \vci{.231^{\diamond}}{.144,.345} & \vci{.183^{\diamond}}{.109,.280} & \vci{.052^{\diamond}}{.046,.059} & \vci{.037^{\diamond}}{.032,.041} \\
\addlinespace[2pt]
DbC (ours) & \vcib{.376^{*}}{.328,.423} & \vcib{.311^{*}}{.271,.353} & \vcib{.394^{*}}{.358,.433} & \vcib{.321^{*}}{.292,.354} & \vcib{.165^{*}}{.143,.192} & \vcib{.131^{*}}{.112,.154} & \vcib{.332^{*}}{.315,.351} & \vcib{.266^{*}}{.252,.282} & \vcib{.265^{*}}{.182,.375} & \vcib{.212^{*}}{.140,.305} & \vcib{.059^{*}}{.053,.065} & \vcib{.042^{*}}{.037,.046} \\
\midrule
\emph{Detection} & AUROC & AP & AUROC & AP & AUROC & AP & AUROC & AP & AUROC & AP & AUROC & AP \\
\midrule
SegVol & \vci{.750}{.728,.769} & \vci{.523}{.487,.562} & \vci{.835}{.816,.854} & \vci{.433}{.395,.474} & \vci{.695}{.669,.720} & \vci{.412}{.378,.452} & \vci{.796}{.779,.812} & \vci{.792}{.766,.816} & \vci{.789}{.750,.822} & \vci{.126}{.097,.161} & \vci{.730}{.711,.748} & \vci{.729}{.702,.755} \\
SAT & \vci{.829}{.811,.846} & \vcib{.598}{.559,.637} & \vci{.861}{.844,.876} & \vci{.494}{.450,.538} & \vci{.809}{.789,.828} & \vci{.512}{.474,.552} & \vci{.837}{.821,.853} & \vci{.731}{.707,.757} & \vci{.738}{.699,.772} & \vci{.098}{.076,.126} & \vci{.794}{.777,.810} & \vci{.783}{.758,.806} \\
VoxTell & \vci{.812}{.793,.831} & \vci{.529}{.499,.564} & \vci{.872}{.855,.887} & \vci{.506}{.462,.554} & \vci{.723}{.701,.746} & \vci{.405}{.374,.438} & \vci{.813}{.797,.831} & \vci{.785}{.758,.812} & \vci{.783}{.748,.813} & \vci{.116}{.092,.146} & \vci{.769}{.751,.787} & \vci{.750}{.724,.776} \\
Disease-VL$^{\dagger}$ & \vci{.785}{.766,.804} & \vci{.530}{.497,.568} & \vci{.872}{.855,.888} & \vci{.517}{.472,.567} & \vci{.825}{.806,.843} & \vci{.565}{.525,.608} & \vci{.828}{.813,.844} & \vcib{.823}{.800,.845} & \vci{.706}{.667,.740} & \vci{.141}{.102,.192} & \vci{.726}{.708,.744} & \vci{.724}{.698,.750} \\
\addlinespace[3pt]
\twolines{DbC, required}{findings only} & \vci{.813^{\diamond}}{.789,.835} & \vci{.584}{.542,.628} & \vci{.916^{\diamond}}{.899,.934} & \vci{.759}{.705,.809} & \vci{.834^{\diamond}}{.810,.855} & \vci{.623}{.570,.675} & \vci{.830^{\diamond}}{.809,.850} & \vci{.804}{.776,.832} & \vci{.761^{\diamond}}{.715,.804} & \vci{.158}{.109,.228} & \vci{.791^{\diamond}}{.769,.813} & \vcib{.800}{.771,.827} \\
\addlinespace[2pt]
DbC (ours) & \vcib{.834}{.817,.851} & \vci{.577}{.542,.614} & \vcib{.932^{*}}{.920,.944} & \vcib{.767^{*}}{.720,.810} & \vcib{.846^{*}}{.828,.862} & \vcib{.624}{.579,.667} & \vcib{.839}{.823,.855} & \vci{.801}{.777,.825} & \vcib{.801}{.764,.837} & \vcib{.165}{.118,.231} & \vcib{.809^{*}}{.793,.825} & \vci{.798}{.774,.821} \\
\bottomrule
\end{tabular}
}
\end{table}

\paragraph{Guideline composition versus a learned layer.}
While DbC uses a fixed, guideline-defined composition, a standard CBM learns a linear classifier over concept predictions. Replacing our parameter-free soft minimum with a learned linear layer over the 40 finding maps yields no improvement on any of the four metrics on seen diseases, while rendering the model unable to predict unseen diseases. The layer is trained on the disease masks, so it has an output only for the diseases it was trained on. The guideline-defined composition therefore preserves auditability and extensibility to new diseases at no cost to accuracy.

\flushfloats
\subsection{Tracing a diagnosis to its findings}
\label{sec:tracing}

Because each diagnosis in DbC is a fixed function of named finding maps, any prediction can be inspected finding by finding. However, such traceability is only as reliable as the finding maps themselves. We therefore score each finding map on its own, on the token grid. Localization is measured by the average precision over tokens (token AP), in which the positive tokens are those that carry the finding and the chance level is their share of all tokens. Detection is measured by the AUROC over CTs, with the maximum of the finding map as the score of a CT and a CT counted as positive when one of its lesions carries the finding. A finding tied to a contrast phase is scored on the CTs where that phase was acquired, as in training.

Table~\ref{tab:findings} reports the model trained on the full training data on the internal test set, as the mean over the findings of three groups. The first two groups hold the required findings of the six diseases, namely the findings that several compositions share and the findings specific to one disease (circles in Table~\ref{tab:compositions}). The third group holds findings that are labeled only from reports or by the size of the lesion, limited to those with at least ten positive test CTs. All three groups are detected reliably and localized far above chance. For the required findings, the token-level average precision reaches $0.715$ across the six findings shared by multiple diagnostic compositions and $0.511$ across the 14 findings specific to a single disease, compared to chance levels of $0.002$ and $0.001$. The shared findings are particularly informative because no single disease label can reproduce them. Rim arterial enhancement, for example, is supervised jointly across lesions of both ICC and metastasis. The third group is detected at a level similar to the required findings, although its labels do not follow from the disease of a lesion. Its token AP is lower, as expected for report findings, which are much rarer than the required ones, and for the small lesions of the two smallest size findings. To verify the ground-truth finding labels used in this evaluation, a senior radiologist visually inspected 200 randomly sampled test lesions against their CT images and confirmed that every assigned required finding was present.

\begin{table}[!htb]
\caption{Accuracy of the finding maps of the model trained on the full training data, on the internal test set, as the mean over the findings of each group. The 95\% confidence interval is under each value.}
\label{tab:findings}
\centering
\tablefont
\setlength{\tabcolsep}{7pt}
\begin{tabular}{@{}lcccc@{}}
\toprule
Group & Findings & Token AP & Chance & AUROC \\
\midrule
Required findings, shared & $6$ & \vci{.715}{.656,.775} & $.002$ & \vci{.924}{.910,.937} \\
Required findings, specific & $14$ & \vci{.511}{.461,.563} & $.001$ & \vci{.919}{.907,.931} \\
Supportive and size findings & $10$ & \vci{.249}{.228,.277} & $<.001$ & \vci{.915}{.903,.927} \\
\bottomrule
\end{tabular}

\end{table}

\begin{figure}[!htb]
\centering
\includegraphics[width=0.82\textwidth]{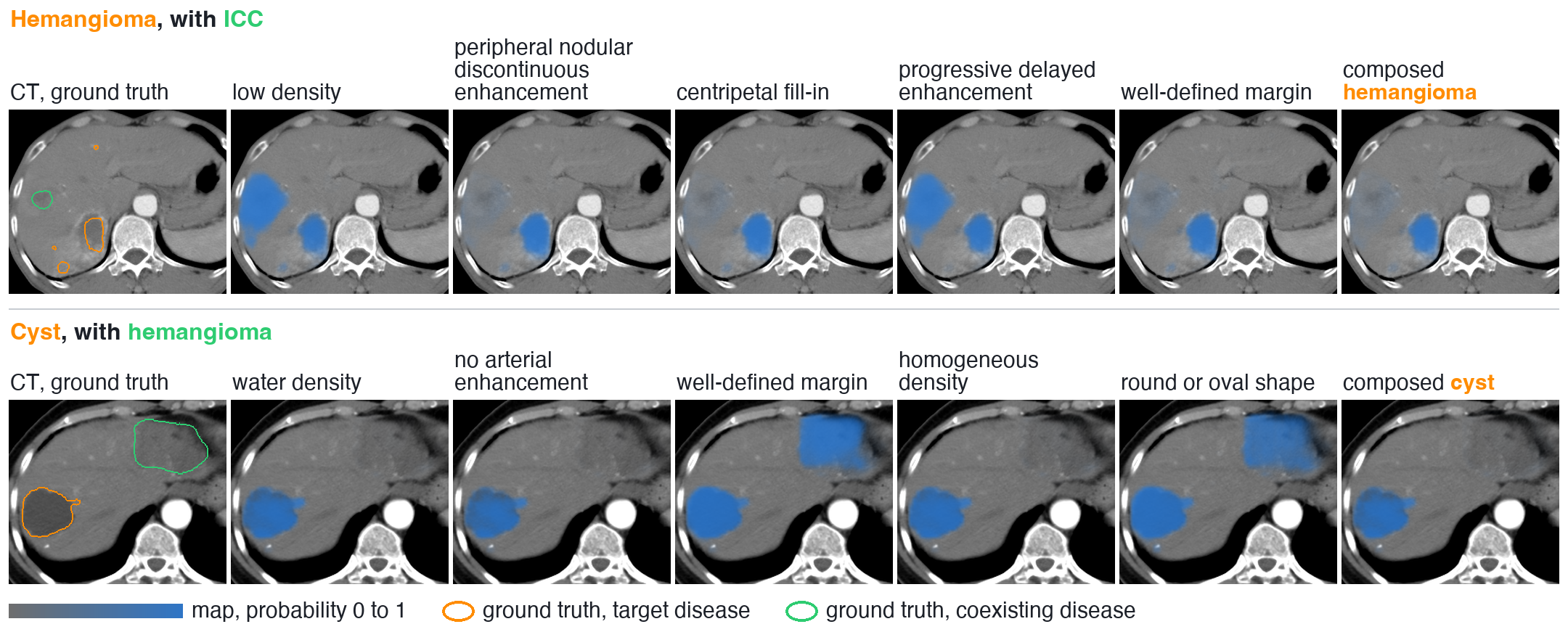}
\caption{Qualitative cases of composition on two CTs that each hold lesions of two diseases from the test set. A single finding can be active on both lesions, as low density and well-defined margin are, whereas the diagnosis map is high only where every required finding is. Appendix~\ref{app:cases} shows further cases.}
\label{fig:qualitative}
\end{figure}

Figure~\ref{fig:qualitative} illustrates this diagnostic traceability on two CT scans that each contain coexisting lesions of two different diseases. While certain required findings are active on both lesions (such as low density in the first scan, and well-defined margin and round-to-oval shape in the second), other required findings activate only on the target lesion. Because the soft minimum yields a high score only where all required findings are present, the coexisting lesion is excluded from the target diagnosis, allowing the reader to see exactly which findings differentiate the two lesions.

\flushfloats
\section{Discussion and conclusion}
\label{sec:discussion}

Our results reflect the clinical order built into DbC. Because every labeled lesion supervises findings shared across multiple diseases, the same training data inform many diagnoses simultaneously. Consequently, DbC outperforms its disease-first counterpart across all training-set scales, with the largest relative gain when data are scarce. Just as a radiologist recognizes an unfamiliar disease as a new combination of familiar findings, DbC composes unseen or rare diseases from findings learned on seen diseases, localizing them significantly better than all comparators. Moreover, because findings describe directly visible properties that recur across diseases and institutions, the learned representation transfers effectively to external cohorts. From a representation learning perspective, these results demonstrate that a vocabulary of visual findings generalizes substantially better than a set of disease labels.

A findings-first model is intrinsically interpretable because its explanation is the computation itself, a diagnosis composed directly from named finding maps that a radiologist can inspect finding by finding. Beyond standalone diagnosis, the representation learned by DbC could also serve as the vision encoder for 3D multimodal large language models (LLMs) that generate radiology reports or answer clinical questions on CT \citep{hamamci2024ctrate, blankemeier2026merlin}, using the finding maps as spatially grounded intermediates. A report states the findings before the impression they support, the order in which DbC reaches a diagnosis, so an LLM conditioned on the finding maps could generate clinical reports with localized finding maps, which will likely make the reports more visually faithful and verifiable. LLM reasoning could likewise build on the finding maps, an intermediate more granular than a disease label and grounded in the image, and we leave these extensions to future work.

This study has three main limitations. First, our evaluation focuses on liver CT, with training data from a single institution covering six common diseases. However, the framework itself is not organ-specific. Because radiological diagnosis across organs is broadly grounded in clinical guidelines, DbC can be extended to other anatomies and pathologies simply by defining the corresponding finding vocabulary and compositions without altering the network architecture. Second, each diagnostic composition currently requires all of its findings to be present, capturing the classic presentation of a disease. While combining guideline and report supervision already trains a diverse vocabulary, extending the composition operator to support richer logical rules (such as disjunctions over alternative finding sets) could better accommodate atypical disease presentations. Third, findings are supervised at the level of whole lesions rather than sub-lesion regions, and a finding required by only a single disease has few labeled examples when that disease is withheld under LODO. Sub-lesion annotations and more comprehensive structured reporting could respectively address these two constraints.

Radiologists name what they see before they name the disease, and published clinical guidelines define each disease by what is visible. Diagnosis by Composition makes these visible findings the fundamental unit of learning. The resulting model is traceable by design, more accurate on seen diseases, and more generalizable to unseen ones.

\bibliographystyle{plainnat}
\bibliography{references}

@incollection{hatamizadeh2022swinunetr,
  author    = {Hatamizadeh, Ali and Nath, Vishwesh and Tang, Yucheng and Yang, Dong and Roth, Holger R. and Xu, Daguang},
  title     = {{Swin UNETR}: Swin Transformers for Semantic Segmentation of Brain Tumors in {MRI} Images},
  booktitle = {Brainlesion: Glioma, Multiple Sclerosis, Stroke and Traumatic Brain Injuries (BrainLes 2021)},
  series    = {Lecture Notes in Computer Science},
  volume    = {12962},
  pages     = {272--284},
  publisher = {Springer},
  year      = {2022},
  doi       = {10.1007/978-3-031-08999-2_22}
}

@inproceedings{he2023swinunetrv2,
  author    = {He, Yufan and Nath, Vishwesh and Yang, Dong and Tang, Yucheng and Myronenko, Andriy and Xu, Daguang},
  title     = {{SwinUNETR-V2}: Stronger {Swin} Transformers with Stagewise Convolutions for {3D} Medical Image Segmentation},
  booktitle = {Medical Image Computing and Computer Assisted Intervention (MICCAI)},
  series    = {Lecture Notes in Computer Science},
  volume    = {14223},
  pages     = {416--426},
  publisher = {Springer},
  year      = {2023},
  doi       = {10.1007/978-3-031-43901-8_40}
}

@article{zhang2023biomedclip,
  author  = {Zhang, Sheng and Xu, Yanbo and Usuyama, Naoto and Xu, Hanwen and Bagga, Jaspreet and Tinn, Robert and Preston, Sam and Rao, Rajesh and Wei, Mu and Valluri, Naveen and others},
  title   = {A Multimodal Biomedical Foundation Model Trained from Fifteen Million Image--Text Pairs},
  journal = {NEJM AI},
  volume  = {2},
  number  = {1},
  pages   = {AIoa2400640},
  year    = {2025},
  doi     = {10.1056/AIoa2400640}
}

@inproceedings{du2024segvol,
  author    = {Du, Yuxin and Bai, Fan and Huang, Tiejun and Zhao, Bo},
  title     = {{SegVol}: Universal and Interactive Volumetric Medical Image Segmentation},
  booktitle = {Advances in Neural Information Processing Systems (NeurIPS)},
  volume    = {37},
  pages     = {110746--110783},
  year      = {2024},
  doi       = {10.52202/079017-3516}
}

@article{zhao2024sat,
  author  = {Zhao, Ziheng and Zhang, Yao and Wu, Chaoyi and Zhang, Xiaoman and Zhou, Xiao and Zhang, Ya and Wang, Yanfeng and Xie, Weidi},
  title   = {Large-Vocabulary Segmentation for Medical Images with Text Prompts},
  journal = {npj Digital Medicine},
  volume  = {8},
  number  = {1},
  pages   = {566},
  year    = {2025},
  doi     = {10.1038/s41746-025-01964-w}
}

@inproceedings{rokuss2025voxtell,
  author    = {Rokuss, Maximilian and Langenberg, Moritz and Kirchhoff, Yannick and Isensee, Fabian and Hamm, Benjamin and Ulrich, Constantin and Regnery, Sebastian and Bauer, Lukas and Katsigiannopulos, Efthimios and Norajitra, Tobias and Maier-Hein, Klaus},
  title     = {{VoxTell}: Free-Text Promptable Universal {3D} Medical Image Segmentation},
  booktitle = {Proceedings of the IEEE/CVF Conference on Computer Vision and Pattern Recognition (CVPR)},
  pages     = {37538--37557},
  year      = {2026}
}

@inproceedings{li2022language,
  author    = {Li, Boyi and Weinberger, Kilian Q. and Belongie, Serge and Koltun, Vladlen and Ranftl, Ren{\'e}},
  title     = {Language-driven Semantic Segmentation},
  booktitle = {International Conference on Learning Representations (ICLR)},
  year      = {2022}
}

@inproceedings{lin2017focal,
  author    = {Lin, Tsung-Yi and Goyal, Priya and Girshick, Ross and He, Kaiming and Doll{\'a}r, Piotr},
  title     = {Focal Loss for Dense Object Detection},
  booktitle = {Proceedings of the IEEE International Conference on Computer Vision (ICCV)},
  pages     = {2980--2988},
  year      = {2017}
}

@inproceedings{milletari2016vnet,
  author    = {Milletari, Fausto and Navab, Nassir and Ahmadi, Seyed-Ahmad},
  title     = {{V-Net}: Fully Convolutional Neural Networks for Volumetric Medical Image Segmentation},
  booktitle = {Fourth International Conference on 3D Vision (3DV)},
  pages     = {565--571},
  year      = {2016},
  doi       = {10.1109/3DV.2016.79}
}

@article{cardoso2022monai,
  author  = {Cardoso, M. Jorge and Li, Wenqi and Brown, Richard and Ma, Nic and Kerfoot, Eric and Wang, Yiheng and Murrey, Benjamin and Myronenko, Andriy and Zhao, Can and Yang, Dong and others},
  title   = {{MONAI}: An open-source framework for deep learning in healthcare},
  journal = {arXiv preprint arXiv:2211.02701},
  year    = {2022}
}

@inproceedings{loshchilov2019adamw,
  author    = {Loshchilov, Ilya and Hutter, Frank},
  title     = {Decoupled Weight Decay Regularization},
  booktitle = {International Conference on Learning Representations (ICLR)},
  year      = {2019}
}

@misc{qwen2025qwen3max,
  author       = {{Qwen Team}},
  title        = {{Qwen3-Max}: Just Scale it},
  howpublished = {Blog post},
  year         = {2025}
}

@article{chernyak2018lirads,
  author  = {Chernyak, Victoria and Fowler, Kathryn J. and Kamaya, Aya and Kielar, Ania Z. and Elsayes, Khaled M. and Bashir, Mustafa R. and Kono, Yuko and Do, Richard K. and Mitchell, Donald G. and Singal, Amit G. and Tang, An and Sirlin, Claude B.},
  title   = {{Liver Imaging Reporting and Data System} ({LI-RADS}) Version 2018: Imaging of Hepatocellular Carcinoma in At-Risk Patients},
  journal = {Radiology},
  volume  = {289},
  number  = {3},
  pages   = {816--830},
  year    = {2018},
  doi     = {10.1148/radiol.2018181494}
}

@inproceedings{mok2020lapirn,
  author    = {Mok, Tony C. W. and Chung, Albert C. S.},
  title     = {Large Deformation Diffeomorphic Image Registration with {Laplacian} Pyramid Networks},
  booktitle = {Medical Image Computing and Computer Assisted Intervention (MICCAI)},
  series    = {Lecture Notes in Computer Science},
  volume    = {12263},
  pages     = {211--221},
  publisher = {Springer},
  year      = {2020},
  doi       = {10.1007/978-3-030-59716-0_21}
}

@article{heinrich2012mind,
  author  = {Heinrich, Mattias P. and Jenkinson, Mark and Bhushan, Manav and Matin, Tahreema and Gleeson, Fergus V. and Brady, Michael and Schnabel, Julia A.},
  title   = {{MIND}: Modality independent neighbourhood descriptor for multi-modal deformable registration},
  journal = {Medical Image Analysis},
  volume  = {16},
  number  = {7},
  pages   = {1423--1435},
  year    = {2012},
  doi     = {10.1016/j.media.2012.05.008}
}

@article{klein2010elastix,
  author  = {Klein, Stefan and Staring, Marius and Murphy, Keelin and Viergever, Max A. and Pluim, Josien P. W.},
  title   = {elastix: A Toolbox for Intensity-Based Medical Image Registration},
  journal = {IEEE Transactions on Medical Imaging},
  volume  = {29},
  number  = {1},
  pages   = {196--205},
  year    = {2010},
  doi     = {10.1109/TMI.2009.2035616}
}

@book{efron1993bootstrap,
  author    = {Efron, Bradley and Tibshirani, Robert J.},
  title     = {An Introduction to the Bootstrap},
  publisher = {Chapman and Hall},
  address   = {New York},
  year      = {1993}
}

@article{yasaka2018liver,
  author  = {Yasaka, Koichiro and Akai, Hiroyuki and Abe, Osamu and Kiryu, Shigeru},
  title   = {Deep Learning with Convolutional Neural Network for Differentiation of Liver Masses at Dynamic Contrast-enhanced {CT}: A Preliminary Study},
  journal = {Radiology},
  volume  = {286},
  number  = {3},
  pages   = {887--896},
  year    = {2018},
  doi     = {10.1148/radiol.2017170706}
}

@article{hamm2019liver,
  author  = {Hamm, Charlie A. and Wang, Clinton J. and Savic, Lynn J. and Ferrante, Marc and Schobert, Isabel and Schlachter, Todd and Lin, MingDe and Duncan, James S. and Weinreb, Jeffrey C. and Chapiro, Julius and Letzen, Brian},
  title   = {Deep learning for liver tumor diagnosis part {I}: development of a convolutional neural network classifier for multi-phasic {MRI}},
  journal = {European Radiology},
  volume  = {29},
  number  = {7},
  pages   = {3338--3347},
  year    = {2019},
  doi     = {10.1007/s00330-019-06205-9}
}

@article{wang2019liver,
  author  = {Wang, Clinton J. and Hamm, Charlie A. and Savic, Lynn J. and Ferrante, Marc and Schobert, Isabel and Schlachter, Todd and Lin, MingDe and Weinreb, Jeffrey C. and Duncan, James S. and Chapiro, Julius and Letzen, Brian},
  title   = {Deep learning for liver tumor diagnosis part {II}: convolutional neural network interpretation using radiologic imaging features},
  journal = {European Radiology},
  volume  = {29},
  number  = {7},
  pages   = {3348--3357},
  year    = {2019},
  doi     = {10.1007/s00330-019-06214-8}
}

@inproceedings{radford2021clip,
  author    = {Radford, Alec and Kim, Jong Wook and Hallacy, Chris and Ramesh, Aditya and Goh, Gabriel and Agarwal, Sandhini and Sastry, Girish and Askell, Amanda and Mishkin, Pamela and Clark, Jack and Krueger, Gretchen and Sutskever, Ilya},
  title     = {Learning Transferable Visual Models From Natural Language Supervision},
  booktitle = {Proceedings of the 38th International Conference on Machine Learning (ICML)},
  series    = {Proceedings of Machine Learning Research},
  volume    = {139},
  pages     = {8748--8763},
  year      = {2021}
}

@inproceedings{liu2023clipdriven,
  author    = {Liu, Jie and Zhang, Yixiao and Chen, Jie-Neng and Xiao, Junfei and Lu, Yongyi and Landman, Bennett A. and Yuan, Yixuan and Yuille, Alan and Tang, Yucheng and Zhou, Zongwei},
  title     = {{CLIP}-Driven Universal Model for Organ Segmentation and Tumor Detection},
  booktitle = {Proceedings of the IEEE/CVF International Conference on Computer Vision (ICCV)},
  pages     = {21095--21107},
  year      = {2023},
  doi       = {10.1109/ICCV51070.2023.01934}
}

@inproceedings{koh2020concept,
  author    = {Koh, Pang Wei and Nguyen, Thao and Tang, Yew Siang and Mussmann, Stephen and Pierson, Emma and Kim, Been and Liang, Percy},
  title     = {Concept Bottleneck Models},
  booktitle = {Proceedings of the 37th International Conference on Machine Learning (ICML)},
  series    = {Proceedings of Machine Learning Research},
  volume    = {119},
  pages     = {5338--5348},
  year      = {2020}
}

@inproceedings{oikarinen2023labelfree,
  author    = {Oikarinen, Tuomas and Das, Subhro and Nguyen, Lam M. and Weng, Tsui-Wei},
  title     = {Label-Free Concept Bottleneck Models},
  booktitle = {International Conference on Learning Representations (ICLR)},
  year      = {2023}
}

@inproceedings{yuksekgonul2023posthoc,
  author    = {Yuksekgonul, Mert and Wang, Maggie and Zou, James},
  title     = {Post-hoc Concept Bottleneck Models},
  booktitle = {International Conference on Learning Representations (ICLR)},
  year      = {2023}
}

@inproceedings{yang2023labo,
  author    = {Yang, Yue and Panagopoulou, Artemis and Zhou, Shenghao and Jin, Daniel and Callison-Burch, Chris and Yatskar, Mark},
  title     = {Language in a Bottle: Language Model Guided Concept Bottlenecks for Interpretable Image Classification},
  booktitle = {Proceedings of the IEEE/CVF Conference on Computer Vision and Pattern Recognition (CVPR)},
  pages     = {19187--19197},
  year      = {2023},
  doi       = {10.1109/CVPR52729.2023.01839}
}

@inproceedings{menon2023visual,
  author    = {Menon, Sachit and Vondrick, Carl},
  title     = {Visual Classification via Description from Large Language Models},
  booktitle = {International Conference on Learning Representations (ICLR)},
  year      = {2023}
}

@inproceedings{misra2017composition,
  author    = {Misra, Ishan and Gupta, Abhinav and Hebert, Martial},
  title     = {From Red Wine to Red Tomato: Composition with Context},
  booktitle = {Proceedings of the IEEE Conference on Computer Vision and Pattern Recognition (CVPR)},
  pages     = {1160--1169},
  year      = {2017},
  doi       = {10.1109/CVPR.2017.129}
}

@inproceedings{nagarajan2018attributes,
  author    = {Nagarajan, Tushar and Grauman, Kristen},
  title     = {Attributes as Operators: Factorizing Unseen Attribute-Object Compositions},
  booktitle = {European Conference on Computer Vision (ECCV)},
  series    = {Lecture Notes in Computer Science},
  volume    = {11205},
  pages     = {172--190},
  publisher = {Springer},
  year      = {2018},
  doi       = {10.1007/978-3-030-01246-5_11}
}

@inproceedings{irvin2019chexpert,
  author    = {Irvin, Jeremy and Rajpurkar, Pranav and Ko, Michael and Yu, Yifan and Ciurea-Ilcus, Silviana and Chute, Chris and Marklund, Henrik and Haghgoo, Behzad and Ball, Robyn and Shpanskaya, Katie and Seekins, Jayne and Mong, David A. and Halabi, Safwan S. and Sandberg, Jesse K. and Jones, Ricky and Larson, David B. and Langlotz, Curtis P. and Patel, Bhavik N. and Lungren, Matthew P. and Ng, Andrew Y.},
  title     = {{CheXpert}: A Large Chest Radiograph Dataset with Uncertainty Labels and Expert Comparison},
  booktitle = {Proceedings of the AAAI Conference on Artificial Intelligence},
  volume    = {33},
  pages     = {590--597},
  year      = {2019},
  doi       = {10.1609/aaai.v33i01.3301590}
}

@inproceedings{jain2021radgraph,
  author    = {Jain, Saahil and Agrawal, Ashwin and Saporta, Adriel and Truong, Steven QH and Duong, Du Nguyen and Bui, Tan and Chambon, Pierre and Zhang, Yuhao and Lungren, Matthew P. and Ng, Andrew Y. and Langlotz, Curtis P. and Rajpurkar, Pranav},
  title     = {{RadGraph}: Extracting Clinical Entities and Relations from Radiology Reports},
  booktitle = {Proceedings of the Neural Information Processing Systems Track on Datasets and Benchmarks},
  volume    = {1},
  year      = {2021}
}

@article{adams2023gpt4,
  author  = {Adams, Lisa C. and Truhn, Daniel and Busch, Felix and Kader, Avan and Niehues, Stefan M. and Makowski, Marcus R. and Bressem, Keno K.},
  title   = {Leveraging {GPT-4} for Post Hoc Transformation of Free-text Radiology Reports into Structured Reporting: A Multilingual Feasibility Study},
  journal = {Radiology},
  volume  = {307},
  number  = {4},
  pages   = {e230725},
  year    = {2023},
  doi     = {10.1148/radiol.230725}
}

@article{hamamci2024ctrate,
  author  = {Hamamci, Ibrahim Ethem and Er, Sezgin and Wang, Chenyu and Almas, Furkan and Simsek, Ayse Gulnihan and Esirgun, Sevval Nil and Dogan, Irem and Durugol, Omer Faruk and Hou, Benjamin and Shit, Suprosanna and others},
  title   = {Generalist Foundation Models from a Multimodal Dataset for {3D} Computed Tomography},
  journal = {Nature Biomedical Engineering},
  volume  = {10},
  number  = {8},
  pages   = {1610--1628},
  year    = {2026},
  doi     = {10.1038/s41551-025-01599-y}
}

@article{han2026clear,
  author  = {Han, Tianyu and Wu, Riga and Tian, Yu and Khader, Firas and Adams, Lisa C. and Bressem, Keno K. and Davatzikos, Christos and Kather, Jakob Nikolas and Shen, Li and Mankoff, David A. and Mortani Barbosa, Jr., Eduardo and Truhn, Daniel},
  title   = {{CLEAR}: an auditable foundation model for radiology grounded in clinical concepts},
  journal = {Nature Biomedical Engineering},
  year    = {2026},
  doi     = {10.1038/s41551-026-01741-4}
}

@article{wu2026mctltdiag,
  author  = {Wu, Xiang'an and Su, Haoyang and Hua, Yiwei and Xu, Yali and Wang, Lilong and Wang, Xiaosong and Wang, Shitian and Jin, Bao and Liu, Xiao and Wan, Xueshuai and Sun, Qiang and Wang, Xuan and Du, Shunda},
  title   = {A Multi-phase {CT} Dataset for Automated Differential Diagnosis of Liver Tumors},
  journal = {Scientific Data},
  volume  = {13},
  pages   = {31},
  year    = {2026},
  doi     = {10.1038/s41597-025-06343-4}
}

@article{blankemeier2026merlin,
  author  = {Blankemeier, Louis and Kumar, Ashwin and Cohen, Joseph Paul and Liu, Jiaming and Liu, Longchao and Van Veen, Dave and Gardezi, Syed Jamal Safdar and Yu, Hongkun and Paschali, Magdalini and Chen, Zhihong and others},
  title   = {{Merlin}: A Computed Tomography Vision--Language Foundation Model and Dataset},
  journal = {Nature},
  volume  = {652},
  number  = {8112},
  pages   = {1318--1328},
  year    = {2026},
  doi     = {10.1038/s41586-026-10181-8}
}

@article{tiu2022chexzero,
  author  = {Tiu, Ekin and Talius, Ellie and Patel, Pujan and Langlotz, Curtis P. and Ng, Andrew Y. and Rajpurkar, Pranav},
  title   = {Expert-level Detection of Pathologies from Unannotated Chest {X}-ray Images via Self-supervised Learning},
  journal = {Nature Biomedical Engineering},
  volume  = {6},
  number  = {12},
  pages   = {1399--1406},
  year    = {2022},
  doi     = {10.1038/s41551-022-00936-9}
}

@inproceedings{huang2021gloria,
  author    = {Huang, Shih-Cheng and Shen, Liyue and Lungren, Matthew P. and Yeung, Serena},
  title     = {{GLoRIA}: A Multimodal Global-Local Representation Learning Framework for Label-efficient Medical Image Recognition},
  booktitle = {Proceedings of the IEEE/CVF International Conference on Computer Vision (ICCV)},
  pages     = {3922--3931},
  year      = {2021},
  doi       = {10.1109/ICCV48922.2021.00391}
}

@inproceedings{wang2022mgca,
  author    = {Wang, Fuying and Zhou, Yuyin and Wang, Shujun and Vardhanabhuti, Varut and Yu, Lequan},
  title     = {Multi-Granularity Cross-modal Alignment for Generalized Medical Visual Representation Learning},
  booktitle = {Advances in Neural Information Processing Systems (NeurIPS)},
  volume    = {35},
  pages     = {33536--33549},
  year      = {2022},
  doi       = {10.52202/068431-2430}
}

@inproceedings{wu2023medklip,
  author    = {Wu, Chaoyi and Zhang, Xiaoman and Zhang, Ya and Wang, Yanfeng and Xie, Weidi},
  title     = {{MedKLIP}: Medical Knowledge Enhanced Language-Image Pre-Training for {X}-ray Diagnosis},
  booktitle = {Proceedings of the IEEE/CVF International Conference on Computer Vision (ICCV)},
  pages     = {21315--21326},
  year      = {2023},
  doi       = {10.1109/ICCV51070.2023.01954}
}

@inproceedings{pellegrini2023xplainer,
  author    = {Pellegrini, Chantal and Keicher, Matthias and {\"O}zsoy, Ege and Jiraskova, Petra and Braren, Rickmer and Navab, Nassir},
  title     = {{Xplainer}: From {X}-Ray Observations to Explainable Zero-Shot Diagnosis},
  booktitle = {Medical Image Computing and Computer Assisted Intervention (MICCAI)},
  series    = {Lecture Notes in Computer Science},
  pages     = {420--429},
  publisher = {Springer},
  year      = {2023},
  doi       = {10.1007/978-3-031-43904-9_41}
}

@article{esr2011reporting,
  author  = {{European Society of Radiology}},
  title   = {Good practice for radiological reporting. {Guidelines} from the {European Society of Radiology (ESR)}},
  journal = {Insights into Imaging},
  volume  = {2},
  number  = {2},
  pages   = {93--96},
  year    = {2011},
  doi     = {10.1007/s13244-011-0066-7}
}

@misc{acr2025communication,
  author       = {{American College of Radiology}},
  title        = {{ACR} Practice Parameter for Communication of Diagnostic Imaging Findings},
  howpublished = {Practice parameter, revised 2025 (Resolution 9)},
  year         = {2025}
}

@article{hartung2020report,
  author  = {Hartung, Michael P. and Bickle, Ian C. and Gaillard, Frank and Kanne, Jeffrey P.},
  title   = {How to Create a Great Radiology Report},
  journal = {RadioGraphics},
  volume  = {40},
  number  = {6},
  pages   = {1658--1670},
  year    = {2020},
  doi     = {10.1148/rg.2020200020}
}

@misc{acgme2019drmilestones,
  author       = {{Accreditation Council for Graduate Medical Education}},
  title        = {Diagnostic Radiology Milestones},
  howpublished = {Second revision, version 2.2},
  year         = {2019}
}

@article{galle2018easl,
  author  = {Galle, Peter R. and Forner, Alejandro and Llovet, Josep M. and Mazzaferro, Vincenzo and Piscaglia, Fabio and Raoul, Jean-Luc and Schirmacher, Peter and Vilgrain, Val{\'e}rie},
  title   = {{EASL} Clinical Practice Guidelines: Management of Hepatocellular Carcinoma},
  journal = {Journal of Hepatology},
  volume  = {69},
  number  = {1},
  pages   = {182--236},
  year    = {2018},
  doi     = {10.1016/j.jhep.2018.03.019}
}

@article{marrero2018aasld,
  author  = {Marrero, Jorge A. and Kulik, Laura M. and Sirlin, Claude B. and Zhu, Andrew X. and Finn, Richard S. and Abecassis, Michael M. and Roberts, Lewis R. and Heimbach, Julie K.},
  title   = {Diagnosis, Staging, and Management of Hepatocellular Carcinoma: 2018 Practice Guidance by the {American Association for the Study of Liver Diseases}},
  journal = {Hepatology},
  volume  = {68},
  number  = {2},
  pages   = {723--750},
  year    = {2018},
  doi     = {10.1002/hep.29913}
}

@inproceedings{shui2025fvlm,
  author    = {Shui, Zhongyi and Zhang, Jianpeng and Cao, Weiwei and Wang, Sinuo and Guo, Ruizhe and Lu, Le and Yang, Lin and Ye, Xianghua and Liang, Tingbo and Zhang, Qi and Zhang, Ling},
  title     = {Large-scale and Fine-grained Vision-language Pre-training for Enhanced {CT} Image Understanding},
  booktitle = {International Conference on Learning Representations (ICLR)},
  year      = {2025}
}

@misc{acr2018liradscore,
  author       = {{American College of Radiology}},
  title        = {{LI-RADS CT/MRI v2018 Core}},
  howpublished = {Liver Imaging Reporting and Data System, version 2018},
  year         = {2018}
}

@article{alvaro2023easlilca,
  author  = {Alvaro, Domenico and Gores, Gregory J. and Walicki, Jo\"el and Hassan, Cesare and Sapisochin, Gonzalo and Komuta, Mina and Forner, Alejandro and Valle, Juan W. and Laghi, Andrea and Ilyas, Sumera I. and Park, Joong-Won and Kelley, Robin K. and Reig, Maria and Sangro, Bruno},
  title   = {{EASL-ILCA} Clinical Practice Guidelines on the management of intrahepatic cholangiocarcinoma},
  journal = {Journal of Hepatology},
  volume  = {79},
  number  = {1},
  pages   = {181--208},
  year    = {2023},
  doi     = {10.1016/j.jhep.2023.03.010}
}

@article{marks2025sarcca,
  author  = {Marks, Robert M. and Arif, Hina and Bali, Maria Antonietta and Brunsing, Ryan L. and Cunha, Guilherme M. and Khasawneh, Hala and El Homsi, Maria and Singh, Charanjeet and Paspulati, Raj and Kierans, Andrea and Qayyum, Aliya},
  title   = {Standardizing the reporting of cholangiocarcinoma: the {Society of Abdominal Radiology} disease-focused panel on cholangiocarcinoma lexicon},
  journal = {Abdominal Radiology},
  volume  = {50},
  number  = {7},
  pages   = {2858--2867},
  year    = {2025},
  doi     = {10.1007/s00261-024-04769-9}
}

@article{kierans2025imaging,
  author  = {Kierans, Andrea S. and Costello, James and Qayyum, Aliya and Taouli, Bachir and Venkatesh, Sudhakar K. and Yoon, Jeong Hee and Bali, Maria A. and Bolan, Candice W. and Lee, Jeong Min and Marks, Robert M. and El Homsi, Maria and Miller, Frank H.},
  title   = {Imaging cholangiocarcinoma: {CT} and {MRI} techniques},
  journal = {Abdominal Radiology},
  volume  = {50},
  number  = {1},
  pages   = {94--108},
  year    = {2025},
  doi     = {10.1007/s00261-024-04216-9}
}

@article{frenette2024acg,
  author  = {Frenette, Catherine and Mendiratta-Lala, Mishal and Salgia, Reena and Wong, Robert J. and Sauer, Bryan G. and Pillai, Anjana},
  title   = {{ACG} Clinical Guideline: Focal Liver Lesions},
  journal = {American Journal of Gastroenterology},
  volume  = {119},
  number  = {7},
  pages   = {1235--1271},
  year    = {2024},
  doi     = {10.14309/ajg.0000000000002857}
}

@article{easl2016benign,
  author  = {{European Association for the Study of the Liver}},
  title   = {{EASL} Clinical Practice Guidelines on the management of benign liver tumours},
  journal = {Journal of Hepatology},
  volume  = {65},
  number  = {2},
  pages   = {386--398},
  year    = {2016},
  doi     = {10.1016/j.jhep.2016.04.001}
}

@article{drenth2022easlcystic,
  author  = {Drenth, Joost and Barten, Thijs and Hartog, Hermien and Nevens, Frederik and Taubert, Richard and Torra Balcells, Roser and Vilgrain, Val\'erie and B\"ottler, Tobias},
  title   = {{EASL} Clinical Practice Guidelines on the management of cystic liver diseases},
  journal = {Journal of Hepatology},
  volume  = {77},
  number  = {4},
  pages   = {1083--1108},
  year    = {2022},
  doi     = {10.1016/j.jhep.2022.06.002}
}

@article{govardhan2026incidental,
  author  = {Govardhan, Chirag and Carney, Benjamin W. and Larson, Michael C. and Lamba, Ramit and Fananapazir, Ghaneh and Corwin, Michael T.},
  title   = {Update on Management of Incidental Findings Seen on Imaging Studies of the Abdomen and Pelvis},
  journal = {RadioGraphics},
  volume  = {46},
  number  = {4},
  pages   = {e250104},
  year    = {2026},
  doi     = {10.1148/rg.250104}
}

@inproceedings{lampert2009attribute,
  author    = {Lampert, Christoph H. and Nickisch, Hannes and Harmeling, Stefan},
  title     = {Learning to Detect Unseen Object Classes by Between-Class Attribute Transfer},
  booktitle = {IEEE Conference on Computer Vision and Pattern Recognition (CVPR)},
  pages     = {951--958},
  year      = {2009},
  doi       = {10.1109/CVPR.2009.5206594}
}

@article{xian2019zeroshot,
  author  = {Xian, Yongqin and Lampert, Christoph H. and Schiele, Bernt and Akata, Zeynep},
  title   = {Zero-Shot Learning---A Comprehensive Evaluation of the Good, the Bad and the Ugly},
  journal = {IEEE Transactions on Pattern Analysis and Machine Intelligence},
  volume  = {41},
  number  = {9},
  pages   = {2251--2265},
  year    = {2019},
  doi     = {10.1109/TPAMI.2018.2857768}
}

@article{zhang2024attribute,
  author  = {Zhang, Runtong and Meng, Fanman and Chen, Shuai and Wu, Qingbo and Xu, Linfeng and Li, Hongliang},
  title   = {Advancing Zero-Shot Semantic Segmentation through Attribute Correlations},
  journal = {Neurocomputing},
  volume  = {594},
  pages   = {127829},
  year    = {2024},
  doi     = {10.1016/j.neucom.2024.127829}
}

@inproceedings{jiang2025malenia,
  author    = {Jiang, Yankai and Lei, Wenhui and Zhang, Xiaofan and Zhang, Shaoting},
  title     = {Unleashing the Potential of Vision-Language Pre-Training for {3D} Zero-Shot Lesion Segmentation via Mask-Attribute Alignment},
  booktitle = {International Conference on Learning Representations (ICLR)},
  year      = {2025}
}

@article{lai2025brgsa,
  author  = {Lai, Haoran and Jiang, Zihang and Yao, Qingsong and Wang, Rongsheng and He, Zhiyang and Tao, Xiaodong and Lv, Weifu and Wei, Wei and Zhou, Shaohua Kevin},
  title   = {Bridged Semantic Alignment for Zero-Shot {3D} Medical Image Diagnosis},
  journal = {IEEE Journal of Biomedical and Health Informatics},
  year    = {2025},
  doi     = {10.1109/JBHI.2025.3629096}
}

@inproceedings{park2025radzero3d,
  author    = {Park, Jonggwon and Choi, Kyoyun and Yoon, Byungmu and Cho, Hong Geun and Hwang, Bumcheol},
  title     = {{RadZero3D}: Bridging Self-Supervised Video Models and Medical Vision-Language Alignment for Zero-Shot Chest {CT} Interpretation},
  booktitle = {IEEE/CVF International Conference on Computer Vision Workshops (ICCVW)},
  pages     = {6801--6808},
  year      = {2025},
  doi       = {10.1109/ICCVW69036.2025.00702}
}

@article{lin2025cxrlt,
  author  = {Lin, Mingquan and Holste, Gregory and Wang, Song and Zhou, Yiliang and Wei, Yishu and Banerjee, Imon and Chen, Pengyi and Dai, Tianjie and Du, Yuexi and Dvornek, Nicha C. and others},
  title   = {{CXR-LT} 2024: A {MICCAI} Challenge on Long-Tailed, Multi-Label, and Zero-Shot Disease Classification from Chest {X}-ray},
  journal = {Medical Image Analysis},
  volume  = {106},
  pages   = {103739},
  year    = {2025},
  doi     = {10.1016/j.media.2025.103739}
}

@article{li2026cdclip,
  author  = {Li, Zheng and Zheng, Xiangwei and Su, Dejian and Zhang, Mingzhe},
  title   = {{CDCLIP}: An Interpretable Zero-Shot Medical Image Classification Framework Based on Concept Decomposition},
  journal = {Biomedical Signal Processing and Control},
  volume  = {116},
  pages   = {109585},
  year    = {2026},
  doi     = {10.1016/j.bspc.2026.109585}
}

@article{bilic2023lits,
  author  = {Bilic, Patrick and Christ, Patrick and Li, Hongwei Bran and Vorontsov, Eugene and Ben-Cohen, Avi and Kaissis, Georgios and Szeskin, Adi and Jacobs, Colin and Mamani, Gabriel Efrain Humpire and Chartrand, Gabriel and others},
  title   = {The Liver Tumor Segmentation Benchmark ({LiTS})},
  journal = {Medical Image Analysis},
  volume  = {84},
  pages   = {102680},
  year    = {2023},
  doi     = {10.1016/j.media.2022.102680}
}

@article{vorontsov2019liver,
  author  = {Vorontsov, Eugene and Cerny, Milena and R{\'e}gnier, Philippe and Di Jorio, Lisa and Pal, Christopher J. and Lapointe, R{\'e}al and Vandenbroucke-Menu, Franck and Turcotte, Simon and Kadoury, Samuel and Tang, An},
  title   = {Deep Learning for Automated Segmentation of Liver Lesions at {CT} in Patients with Colorectal Cancer Liver Metastases},
  journal = {Radiology: Artificial Intelligence},
  volume  = {1},
  number  = {2},
  pages   = {180014},
  year    = {2019},
  doi     = {10.1148/ryai.2019180014}
}

@article{rahmatullah2026abscess,
  author  = {Rahmatullah, Zahra F. and Arshad, Hajra and Chu, Linda C. and Kawamoto, Satomi and Fishman, Elliot K.},
  title   = {Hepatic Abscess on {CT}: Pearls, Pitfalls and Diagnostic Challenges},
  journal = {Emergency Radiology},
  volume  = {33},
  number  = {5},
  pages   = {1209--1217},
  year    = {2026},
  doi     = {10.1007/s10140-026-02516-9}
}

@article{lohne2026pretext,
  author  = {Lohne, Siri Margrete Hugo and Hebelka, Hanna and Prince, Jip Frederik and Bohte, Anneloes Elisabeth and Littooij, Annemieke Simone and de Lange, Charlotte and Deganello, Annamaria and Dias, Silvia Costa and Chambers, Greg and Toso, Seema and Habre, C{\'e}line and Pace, Erika and Franchi-Abella, St{\'e}phanie and Woodley, Helen and McGuirk, Simon P.},
  title   = {Challenges with the {PRETEXT} Staging System: A {SIOPEL} {Radiology} {Group} Position Paper},
  journal = {Pediatric Radiology},
  year    = {2026},
  note    = {Online ahead of print},
  doi     = {10.1007/s00247-026-06786-w}
}

\clearpage
\appendix

\section{Finding catalog and diagnostic compositions}
\label{app:catalog}

\begin{figure}[p]
\centering
\includegraphics[width=0.78\textwidth]{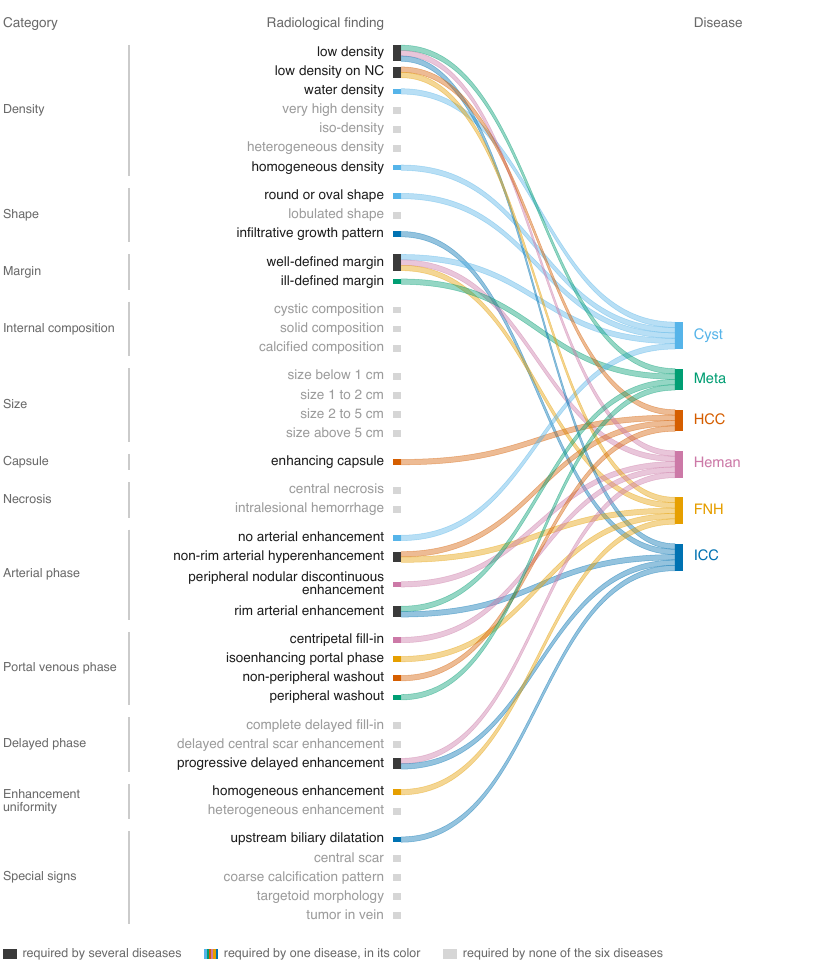}
\caption{The finding vocabulary and the six diagnostic compositions drawn as a flow. The 40 findings are on the left in the order of Table~\ref{tab:catalog}, and the six diseases are on the right (Meta metastasis, Heman hemangioma), ordered to reduce ribbon crossings. A ribbon joins a finding to every disease whose composition requires it. A finding required by several diseases (dark bar) fans out into several ribbons, a finding required by one disease (bar in the color of that disease) has a single ribbon, and a finding required by none of the six (light bar, gray label) has none.}
\label{fig:flow}
\end{figure}

Radiologists on our team who specialize in liver imaging defined the finding catalog and derived each diagnostic composition from published radiological guidelines and imaging references (Section~\ref{sec:vocabulary}). Table~\ref{tab:catalog} lists the 40 findings of the catalog in its twelve categories, each with its definition, the contrast phases on which it is supervised, and the diagnostic compositions that require it. Figure~\ref{fig:flow} draws the same membership as a flow from the findings to the six diseases, and Table~\ref{tab:compositions} writes out every composition.

\begin{table}[p]
\caption{The finding catalog. Phases gives the contrast phases on which a finding is supervised (NC non-contrast, A arterial, PV portal venous, D delayed). Required in lists those of the six diagnostic compositions that contain the finding (Meta metastasis, Heman hemangioma), and the two rare-disease compositions appear in Table~\ref{tab:compositions} only.}
\label{tab:catalog}
\centering
\tablefontsmall
\renewcommand{\arraystretch}{0.9}
\makebox[\textwidth][c]{\begin{tabular}{@{}l@{\hspace{8pt}}p{0.51\textwidth}@{\hspace{8pt}}l@{\hspace{8pt}}l@{}}
\toprule
Finding & Definition & Phases & Required in \\
\midrule
\multicolumn{4}{@{}l}{\emph{Density}} \\
\hspace{6pt}low density & Attenuation below that of the surrounding liver parenchyma. & any & ICC, Meta, Heman \\
\hspace{6pt}low density on NC & Attenuation below that of the liver on the non-contrast phase, whatever it is on the contrast phases. & NC & HCC, FNH \\
\hspace{6pt}water density & Attenuation in the fluid range, about $-10$ to $20$~HU. & any & Cyst \\
\hspace{6pt}very high density & Marked hyperattenuation, typically above $100$~HU. & any &  \\
\hspace{6pt}iso-density & Attenuation similar to that of the surrounding liver parenchyma. & any &  \\
\hspace{6pt}heterogeneous density & Non-uniform internal attenuation, with regions of lower and of higher density. & any &  \\
\hspace{6pt}homogeneous density & Uniform internal attenuation throughout the lesion. & any & Cyst \\
\midrule
\multicolumn{4}{@{}l}{\emph{Shape}} \\
\hspace{6pt}round or oval shape & Spherical or ovoid contour. & any & Cyst \\
\hspace{6pt}lobulated shape & Contour with several lobes, not smoothly round. & any &  \\
\hspace{6pt}infiltrative growth pattern & Extension along tissue planes or vessels without a discrete boundary. & any & ICC \\
\midrule
\multicolumn{4}{@{}l}{\emph{Margin}} \\
\hspace{6pt}well-defined margin & Sharp, well-circumscribed boundary with the surrounding parenchyma. & any & Heman, FNH, Cyst \\
\hspace{6pt}ill-defined margin & Poorly demarcated, fuzzy or irregular boundary. & any & Meta \\
\midrule
\multicolumn{4}{@{}l}{\emph{Internal composition}} \\
\hspace{6pt}cystic composition & Fluid-filled internal structure. & any &  \\
\hspace{6pt}solid composition & Solid internal structure without a fluid component. & any &  \\
\hspace{6pt}calcified composition & Calcium within the lesion. & any &  \\
\midrule
\multicolumn{4}{@{}l}{\emph{Size}} \\
\hspace{6pt}size below 1 cm & Long-axis diameter below 1~cm. & any &  \\
\hspace{6pt}size 1 to 2 cm & Long-axis diameter of at least 1~cm and below 2~cm. & any &  \\
\hspace{6pt}size 2 to 5 cm & Long-axis diameter between 2 and 5~cm. & any &  \\
\hspace{6pt}size above 5 cm & Long-axis diameter above 5~cm. & any &  \\
\midrule
\multicolumn{4}{@{}l}{\emph{Capsule}} \\
\hspace{6pt}enhancing capsule & Peripheral rim that enhances on the portal venous or delayed phase, a fibrous capsule. & PV or D & HCC \\
\midrule
\multicolumn{4}{@{}l}{\emph{Necrosis}} \\
\hspace{6pt}central necrosis & Non-enhancing necrotic center, typically of low attenuation within an enhancing rim. & any &  \\
\hspace{6pt}intralesional hemorrhage & Foci of hemorrhage within the lesion, typically hyperdense on the non-contrast phase. & any &  \\
\midrule
\multicolumn{4}{@{}l}{\emph{Arterial phase}} \\
\hspace{6pt}no arterial enhancement & No enhancement on the arterial phase, at the low attenuation of the non-contrast phase. & A & Cyst \\
\hspace{6pt}non-rim arterial hyperenhancement & Arterial enhancement that is not rim-like and unequivocally greater than that of the liver, in all or part of the lesion. & A & HCC, FNH \\
\hspace{6pt}peripheral nodular discontinuous enhancement & Discontinuous nodular foci of arterial enhancement along the periphery of the lesion. & A & Heman \\
\hspace{6pt}rim arterial enhancement & Continuous peripheral ring of arterial enhancement around a less enhancing center. & A & ICC, Meta \\
\midrule
\multicolumn{4}{@{}l}{\emph{Portal venous phase}} \\
\hspace{6pt}centripetal fill-in & Enhancement that extends from the periphery toward the center on the portal venous and later phases. & PV and D & Heman \\
\hspace{6pt}isoenhancing portal phase & Attenuation similar to that of the surrounding liver on the portal venous phase. & PV & FNH \\
\hspace{6pt}non-peripheral washout & Enhancement of the non-peripheral part of the lesion that falls below that of the liver on the portal venous phase. & A and PV & HCC \\
\hspace{6pt}peripheral washout & Enhancement of the peripheral rim of the lesion that falls below that of the liver on the portal venous phase. & PV & Meta \\
\midrule
\multicolumn{4}{@{}l}{\emph{Delayed phase}} \\
\hspace{6pt}complete delayed fill-in & Complete filling with contrast on the delayed phase, at or above the attenuation of the liver. & PV and D &  \\
\hspace{6pt}delayed central scar enhancement & Enhancement of a central scar on the delayed phase. & D &  \\
\hspace{6pt}progressive delayed enhancement & Enhancement that increases from the portal venous to the delayed phase. & PV and D & ICC, Heman \\
\midrule
\multicolumn{4}{@{}l}{\emph{Enhancement uniformity}} \\
\hspace{6pt}homogeneous enhancement & Uniform enhancement throughout the lesion on the phases where it enhances. & A or PV & FNH \\
\hspace{6pt}heterogeneous enhancement & Non-uniform, patchy or mosaic enhancement on the phases where the lesion enhances. & A or PV &  \\
\midrule
\multicolumn{4}{@{}l}{\emph{Special signs}} \\
\hspace{6pt}upstream biliary dilatation & Dilated intrahepatic bile ducts upstream of the lesion. & any & ICC \\
\hspace{6pt}central scar & Central stellate or linear region of low attenuation within the lesion, a fibrous scar. & any &  \\
\hspace{6pt}coarse calcification pattern & Large, irregular or amorphous calcifications. & any &  \\
\hspace{6pt}targetoid morphology & Concentric layers on the contrast phases, typically a hypodense center, an enhancing ring and a hypodense outer ring. & A or PV &  \\
\hspace{6pt}tumor in vein & Extension of the lesion into a vein, often an enhancing filling defect that expands the vessel. & any &  \\
\bottomrule
\end{tabular}
}
\end{table}

\paragraph{Definitions.}
The findings are standard terms of radiological reporting and come from the guidelines and lexicons cited below. Each definition in Table~\ref{tab:catalog} states what a radiologist looks for in the image. A finding label is never computed from the image by a rule such as a threshold on attenuation. A lesion is labeled with a finding when the diagnostic composition of its disease requires it or when its report describes it, and with one of the four size findings according to the dimensions of its mask (Section~\ref{app:labels}).

\paragraph{Contrast phases.}
A finding marked ``any'' in the Phases column has no phase requirement and is supervised on every CT whichever phases were acquired. A finding tied to a contrast phase is supervised only on CTs where that phase was acquired. Where the column joins two phases by ``and'', both must be present, as for the findings that describe a change between phases, and where it joins them by ``or'', either one suffices.

\paragraph{Guideline sources.}
The composition of hepatocellular carcinoma (HCC) is derived from LI-RADS and the EASL and AASLD guidelines \citep{chernyak2018lirads, galle2018easl, marrero2018aasld}, and that of intrahepatic cholangiocarcinoma (ICC) from the EASL-ILCA guideline and society imaging references \citep{alvaro2023easlilca, marks2025sarcca, kierans2025imaging}. The compositions of metastasis, hemangioma, focal nodular hyperplasia (FNH) and cyst are derived from the ACG focal-liver-lesion guideline \citep{frenette2024acg}. For metastasis, the LI-RADS criteria for malignancy other than hepatocellular carcinoma (LR-M) are a further source \citep{acr2018liradscore}. For the three benign diseases, the further sources are the EASL benign-tumor and cystic-disease guidelines and a current imaging review \citep{easl2016benign, drenth2022easlcystic, govardhan2026incidental}. The radiologists on our team defined the compositions of the two rare diseases at evaluation time from published imaging criteria, for abscess \citep{rahmatullah2026abscess} and for hepatoblastoma \citep{lohne2026pretext}, with no reference to the model or its outputs (Section~\ref{sec:beyond}).

\section{Implementation details}
\label{app:details}

\subsection{Data and preprocessing}
\label{app:data}

\paragraph{Data.}
We curated 15,609 multi-phase liver CTs with lesion masks and radiology reports from a single institution (Section~\ref{sec:setup}). The masks label each lesion with one of the lesion classes listed in Section~\ref{app:metrics} or with one of two catch-all classes, other malignant and other benign, which have no composition. Lesions of the two catch-all classes contribute supportive and size findings to the finding maps but no required findings, and CTs without a lesion supply negatives at every token. Table~\ref{tab:cohorts} gives the sizes of the internal splits and of the two external cohorts.

\begin{table}[!htb]
\caption{Cohorts. The internal data are split at the patient level, and localization on the internal test set is scored over its CTs with lesions.}
\label{tab:cohorts}
\centering
\footnotesize
\begin{tabular}{@{}llrr@{}}
\toprule
Cohort & Subset & CTs & Lesions \\
\midrule
\multirow{4}{*}{Internal} & Training & 12{,}487 & \\
 & Validation & 1{,}559 & \\
 & Test & 1{,}563 & \\
 & Test CTs with lesions & 533 & 2{,}571 \\
\midrule
\multirow{5}{*}{MCT-LTDiag} & HCC & 91 & 191 \\
 & ICC & 81 & 286 \\
 & Metastasis & 216 & 1{,}240 \\
 & Hemangioma & 94 & 452 \\
 & All four diseases & 482 & 2{,}169 \\
\midrule
\multirow{3}{*}{\shortstack[l]{Rare diseases,\\different institution}} & Abscess & 176 & \\
 & Hepatoblastoma & 43 & \\
 & Without a liver lesion & 119 & \\
\bottomrule
\end{tabular}

\end{table}

\paragraph{Preprocessing.}
Contrast phases not already aligned to the non-contrast phase are registered to it with a Laplacian pyramid network \citep{mok2020lapirn} driven by modality-independent neighbourhood descriptors \citep{heinrich2012mind}, and a missing phase is stored as an all-zero channel. Volumes are resampled to $0.75\times0.75\times5.0$~mm, clipped to $[-75,250]$~HU, scaled to $[0,1]$ and cropped around the liver to $384\times288\times64$ voxels, and the lesions are the 6-connected components of the mask of each class. The token grid is $96\times72\times16$, so $N=110{,}592$, with $4\times4\times4$ voxels ($3\times3\times20$~mm) per token. The published baselines read the same CTs, split and masks with the wider window $[-500,1000]$~HU that their own normalizers expect.

\paragraph{Split.}
The 15,609 CTs are divided into 12,487 training, 1,559 validation and 1,563 test CTs by a patient-level 80/10/10 split with seed 42, stratified by the rarest disease present in each CT, with a separate stratum for CTs without lesions. Of the 1,563 test CTs, 533 carry lesions, with 2,571 lesions in total, and the localization metrics are computed over these CTs.

\flushfloats
\subsection{Finding labels from masks and reports}
\label{app:labels}

\begin{figure}[!htb]
\centering
\includegraphics[width=\textwidth]{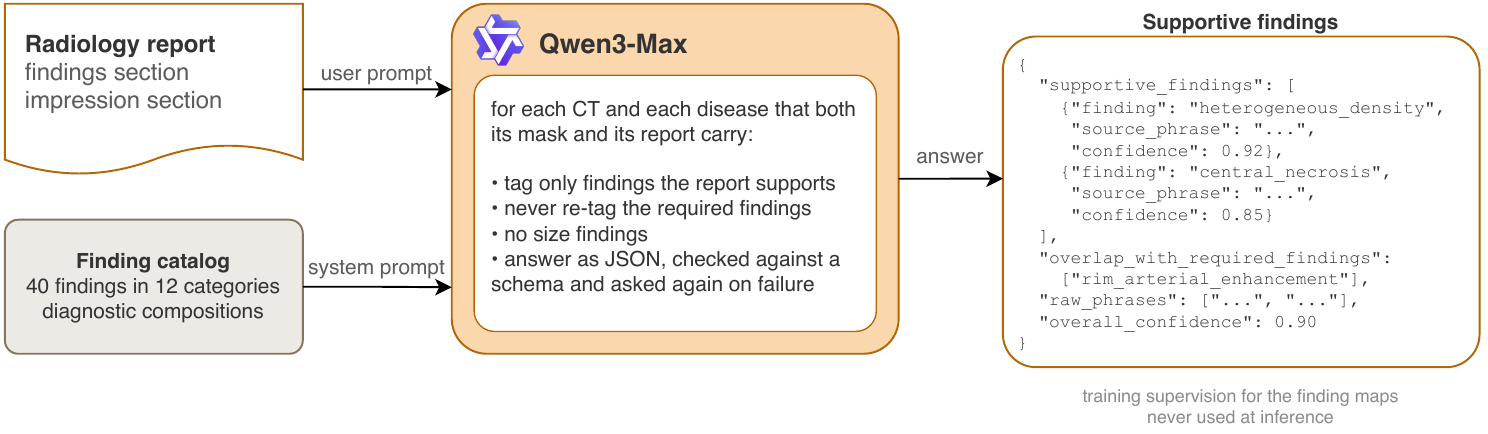}
\caption{Extraction of supportive findings from the report. For every CT and each disease that both its mask and its report carry, Qwen3-Max receives the finding catalog and the diagnostic compositions in its system prompt. The user prompt holds the report sections, the disease and the long axes of the lesions. The model answers in JSON with the supported findings, their source phrases and a confidence (example values shown). The supportive findings supervise the finding maps during training and are never used at inference.}
\label{fig:extraction}
\end{figure}

\paragraph{Supportive findings.}
Figure~\ref{fig:extraction} illustrates the extraction of supportive findings. For every CT and each disease that both its mask and its report carry, Qwen3-Max \citep{qwen2025qwen3max} receives the finding catalog and the diagnostic compositions (Section~\ref{app:catalog}) with the findings and impression sections of the report. The model returns in a fixed schema the catalog findings that the report supports, excluding the required findings of the disease and the size findings, with the report phrase behind each. A radiologist on our team reviewed a random sample of 200 extractions and found all of them faithful to the source report.

\paragraph{Assembly.}
The finding labels of each lesion are assembled deterministically from the required findings of its disease, the supportive findings of its CT and disease, and one size finding derived from the dimensions of its mask. A lesion whose disease the report does not mention keeps its required and size findings only. The finding labels are not mutually exclusive, and every finding map is supervised on its own. When a report describes a lesion with a finding that contradicts a required finding of its disease, a metastasis with a well-defined margin for instance, the lesion keeps its required findings and also receives the report finding as a supportive finding. A report can therefore only add findings to a lesion. Reports that contradict a required finding in this way are uncommon in our data. The extraction never changes a diagnostic composition, which consists of the required findings of its disease alone.

\paragraph{Rasterization.}
The finding labels of every lesion are rasterized to the token grid. For every token and every class with at least one voxel in the token, the token is assigned the lesion of that class that holds the majority of its voxels, and all findings of that lesion are set positive. A token touched by lesions of two classes receives the findings of both. Low density on the non-contrast phase implies low density, a finding tied to a contrast phase is supervised only on CTs where that phase was acquired (Table~\ref{tab:catalog}), and tokens outside the liver are negatives for every finding. The masks of the classes that have a composition, rasterized on the same grid with the same rule, are the training targets of Disease-VL and Disease-V and the reference against which every model is scored.

\flushfloats
\subsection{Training}
\label{app:training}

\paragraph{Shared recipe.}
DbC, Disease-VL and Disease-V are trained with the recipe of Table~\ref{tab:recipe}, and the only difference between the three is the output layer described in Section~\ref{sec:setup}. The hyperparameters were searched separately for Disease-VL and for DbC, and the search chose the same values. Checkpoints are selected by the macro token AP on the validation split, the average precision over tokens averaged over the outputs of a model. Training CTs are drawn with replacement, about half of them with lesions and half without. Each CT with lesions is assigned to the rarest disease it contains and drawn with weight proportional to $q^{-1/2}$, where $q$ is the number of training CTs assigned to that disease, so each disease is drawn in proportion to $\sqrt{q}$ rather than $q$.

\begin{table}[!htb]
\caption{Shared training recipe of DbC, Disease-VL and Disease-V.}
\label{tab:recipe}
\centering
\footnotesize
\begin{tabular}{@{}p{0.27\textwidth}p{0.66\textwidth}@{}}
\toprule
Hyperparameter & Value \\
\midrule
Hardware & 2 NVIDIA H20 GPUs, distributed data parallel \\
Micro-batch and accumulation & 2 CTs per GPU, 8 micro-batches \\
Effective batch & 32 CTs per optimizer step \\
Optimizer & AdamW \citep{loshchilov2019adamw}, peak learning rate $5.7\times10^{-5}$, weight decay 0.1 \\
Schedule & linear warm-up over the first 3\% of optimizer steps, then constant \\
Gradient clipping & global norm 1.0 \\
Precision & float16 mixed precision with gradient scaling \\
Epochs & at most 80, with validation every epoch on at most 128 batches \\
Early stopping & patience 8 validation passes \\
Checkpoint selection & validation macro token AP \\
Loss & Dice-focal with focal $\alpha=0.25$ and $\gamma=2$, Dice and focal weights 1 and 1, Dice smoothing $10^{-5}$ \\
Masks in the loss & contrast-phase mask of each finding, tokens outside the liver as negatives \\
Augmentation and dropout & none, encoder dropout 0 \\
Encoder & SwinUNETR-V2 \citep{he2023swinunetrv2} in the MONAI implementation \citep{cardoso2022monai}, feature size 48, depths (2, 2, 2, 2), heads (3, 6, 12, 24), window size 7 \\
\bottomrule
\end{tabular}
\end{table}

\paragraph{Readout.}
The projection of the image branch in Figure~\ref{fig:fig2} is a learned linear map from the 96-dimensional decoder feature of each token to the 512-dimensional token embedding $v_n$ of Eq.~(\ref{eq:readout}). Each finding is placed in the prompt ``a CT image of a liver lesion showing'' followed by its name as the file \texttt{findings.json} of our code lists it, which words some names at greater length than Table~\ref{tab:catalog} (Section~\ref{app:code}). The adapter $\phi$ consists of two residual blocks, each a layer normalization, a 512-dimensional linear layer, a GELU and a second 512-dimensional linear layer. The adapter needs no input projection, since the text embedding already has 512 dimensions. The scale is parameterized as $s_m=\exp(\gamma_m)$, initialized at $1/0.07$ and capped at 100, and the bias $b_m$ is initialized at the log-odds $\log\big(\pi_m/(1-\pi_m)\big)$ of the prevalence $\pi_m$ of finding $m$ among training tokens, with $\pi_m=0.01$ when the prevalence is unavailable. Disease-VL uses the same projection, adapter and readout with disease names in place of finding names, in the prompt ``a CT image of a liver lesion of type'' followed by the name, and Disease-V initializes the bias of each disease output from its prevalence in the same way.

\paragraph{Leave-one-disease-out.}
Leave-one-disease-out (LODO) trains one model per unseen disease. For an unseen disease $k$, no CT whose mask contains $k$ is used to train the model, and every other CT keeps its split assignment. Each model is trained with its usual recipe. For Disease-VL and the published baselines the output of $k$ is also removed from the loss, and each is prompted with the name of $k$ at test time. The positives of the test cohort of $k$ are CTs that carry $k$ and were not used for training, and its negatives are the CTs of the test split that do not carry $k$.

\paragraph{Data scaling.}
For each run, every stratum of the training split is shuffled once, and a subset keeps a prefix of each stratum, so that smaller subsets are nested in larger ones. Seeds 42, 43 and 44 set both the subset and the initialization, the validation and test splits are unchanged, and all other settings follow Table~\ref{tab:recipe}.

\paragraph{Published baselines.}
Each published baseline starts from its released checkpoint and is fine-tuned on the voxel masks of our training data at native resolution with its published recipe and seed 42, for the same number of optimizer steps as the DbC run of the same setting. Its text encoder and prompt embeddings are frozen, and its checkpoint is selected by validation macro token AP.

\paragraph{Learned composition.}
The learned composition of Table~\ref{tab:analysisseen} keeps the finding maps of the DbC model and replaces the soft minimum with a linear layer, applied at every token, that maps the 40 finding probabilities to one logit for each class that has a composition and is followed by a sigmoid. The layer is trained on the masks of these classes with the loss of Table~\ref{tab:recipe}, and its checkpoint is selected by validation macro token AP like those of the other models.

\paragraph{Required findings only.}
For the row ``DbC, required findings only'' of Tables~\ref{tab:analysisseen} and~\ref{tab:analysisunseen}, the supportive and size findings are removed from the finding labels and DbC is trained on the required findings alone, with the same recipe, on the full training data and on every LODO split. The compositions are unchanged, since they consist of required findings only.

\flushfloats
\subsection{Comparators and prompts}
\label{app:prompts}

\paragraph{Adaptation of the published baselines.}
SegVol, SAT and VoxTell are wrapped so that each returns one full-resolution map per disease from the four-phase input, and the maps are max-pooled to the token grid, since a token of the reference masks is positive when any of its voxels is. The one-channel input layer of each model is extended to the four phases by repeating its weights and dividing by four, which preserves the pretrained response when the phases are identical, and each phase is normalized with the rule of the model. SegVol is used with text prompts only and resized to its fixed input shape, SAT in its released Nano configuration with its BioLORD text embeddings, and VoxTell as the full text-conditioned model with its positional encoding adapted to our crop.

\paragraph{Name prompts.}
The published baselines are prompted with plain clinical names, namely hepatocellular carcinoma, intrahepatic cholangiocarcinoma, liver metastasis, hepatic hemangioma, focal nodular hyperplasia of the liver and hepatic cyst, and abscess and hepatoblastoma for the rare diseases. Disease-VL receives the name of each disease in its own prompt (Section~\ref{app:training}).

\paragraph{Description prompts.}
In the second control of Section~\ref{sec:analysis}, the same trained models are prompted with a description of each disease by the required findings of its diagnostic composition (Table~\ref{tab:compositions}), which are the findings that DbC composes. For HCC, the description given to the published baselines reads ``liver lesion with low density on the non-contrast phase, non-rim arterial hyperenhancement, non-peripheral washout and enhancing capsule'', and Disease-VL receives the same findings after the words ``a CT image of a liver lesion showing''. The weights are unchanged, and each model is evaluated with its checkpoint trained on the full training data and with its six LODO checkpoints.

\flushfloats
\subsection{Metrics and confidence intervals}
\label{app:metrics}

\paragraph{Scores.}
Every model supplies one map per disease on the token grid, namely the composed diagnosis map for DbC, the sigmoid of the disease output for Disease-VL and Disease-V, and the max-pooled sigmoid of the native logits for the published baselines. The score of a CT for a disease is the maximum of its map, and the CT is positive when it holds a lesion of that disease.

\paragraph{Detection.}
AUROC and average precision (AP) are computed over CTs. A disease without positive CTs in a cohort is left out of the macro mean of that cohort.

\paragraph{Localization.}
The predicted lesions are the 6-connected components of the thresholded map. Each lesion of the reference masks takes the Dice and the IoU of the overlapping predicted component with the largest IoU, a lesion without an overlapping component scores zero, and the values are averaged over lesions.

\paragraph{Lesion classes and macro means.}
A macro mean is the unweighted mean over the lesion classes of a cohort, recomputed in every bootstrap replicate. The results for unseen diseases and for rare diseases are reported per disease, and on MCT-LTDiag the classes are its four diseases. On the internal test set the classes are the six diseases. This holds for the macro means of Table~\ref{tab:seen}, Figure~\ref{fig:scaling} and Table~\ref{tab:analysisseen}, and for those of Table~\ref{tab:scaling}.

\paragraph{Bootstrap and runs.}
The bootstrap draws 2,000 patient-level replicates, in which the CTs and lesions of a patient move with the patient and every model receives the same patients, and the 95\% interval is the percentile interval \citep{efron1993bootstrap}. Significance is read from the interval of the paired difference between two models, never from the overlap of their intervals. In data scaling every run is scored on the same test CTs, a curve value is the mean over the three runs, and significance comes from the bootstrap of the mean over runs (Table~\ref{tab:scaling}).

\flushfloats
\subsection{External cohorts}
\label{app:external}

\paragraph{MCT-LTDiag.}
The public source \citep{wu2026mctltdiag} holds four-phase CT studies of five types of liver lesion, one type per study and none below 1~cm, and its colorectal and breast cancer metastases are merged into metastasis. The source registers the non-contrast, arterial and delayed volumes to the portal venous phase with B-spline registration in elastix \citep{klein2010elastix}, and its lesion masks are automated outputs corrected by a senior radiologist. The internal preprocessing (resampling, crop, window, connected components, rasterization) is applied unchanged, without the internal registration step. After an audit of the source files, 482 CTs with 2,169 lesions are used (Table~\ref{tab:cohorts}). MCT-LTDiag is used for testing only, and no weight or prompt was ever fit on it. Every CT of the cohort carries one of the four diseases, so the negatives of a disease are the CTs of the other three diseases, and macro values are over the four diseases.

\paragraph{Rare-disease cohort.}
The CTs of the rare-disease cohort come from a different institution than the internal data, and neither abscess nor hepatoblastoma appears in the internal training, validation or test data. Studies with the contrast phases that the composition of the disease reads (arterial for abscess, arterial and portal venous for hepatoblastoma) are processed with the internal geometry and window and screened against the internal training split on liver extent, liver fraction in the crop and the intensity statistics of each phase. Table~\ref{tab:cohorts} gives the resulting cohort. Every model answers from its checkpoint trained on the full training data. DbC evaluates the compositions of Table~\ref{tab:compositions}, and the other models are prompted with the disease name.

\flushfloats
\section{Additional results}
\label{app:results}

This section gives the results that the main text cites without a table of its own. Localization is reported by Dice and IoU and detection by AUROC and AP (Section~\ref{app:metrics}). Unless a caption says otherwise, the 95\% confidence interval is printed under each value, bold marks the best value and $^{\dagger}$ a matched ablation.

\flushfloats
\subsection{Data scaling}
\label{app:scaling}

Table~\ref{tab:scaling} lists the values plotted in Figure~\ref{fig:scaling}, namely the mean over the three runs at each training-set size with the standard deviation across the runs, and adds the same curves for the malignant subset of HCC, ICC and metastasis. Each value of Table~\ref{tab:scaling} is a mean over three runs, whereas Table~\ref{tab:seen} reports one run, so the two tables differ at the full training-set size.

\begin{table}[!htb]
\caption{Data scaling, the values behind Figure~\ref{fig:scaling} (macro means, Section~\ref{app:metrics}). Each entry is the mean over three runs at one training-set size, with the standard deviation across the runs under it. A star marks a significant advantage of DbC over Disease-VL, from the patient-level bootstrap of the mean over runs, and $^{\dagger}$ a matched ablation.}
\label{tab:scaling}
\centering
\tablefont
\setlength{\tabcolsep}{5pt}
\begin{tabular}[t]{@{}lcccccc@{}}
\toprule
Model & 10\% & 20\% & 40\% & 60\% & 80\% & 100\% \\
\midrule
\multicolumn{7}{@{}l}{\emph{Macro mean, Dice}} \\
DbC (ours) & \vsd{.070^{*}}{.010} & \vsd{.103^{*}}{.011} & \vsd{.164^{*}}{.019} & \vsd{.219^{*}}{.014} & \vsd{.275^{*}}{.019} & \vsd{.312^{*}}{.006} \\
Disease-VL$^{\dagger}$ & \vsd{.054}{.001} & \vsd{.094}{.014} & \vsd{.149}{.010} & \vsd{.208}{.009} & \vsd{.248}{.021} & \vsd{.282}{.010} \\
\midrule
\multicolumn{7}{@{}l}{\emph{Macro mean, AUROC}} \\
DbC (ours) & \vsd{.725^{*}}{.009} & \vsd{.768}{.011} & \vsd{.836^{*}}{.016} & \vsd{.864^{*}}{.022} & \vsd{.898^{*}}{.005} & \vsd{.918^{*}}{.011} \\
Disease-VL$^{\dagger}$ & \vsd{.700}{.006} & \vsd{.758}{.007} & \vsd{.822}{.004} & \vsd{.855}{.013} & \vsd{.882}{.008} & \vsd{.903}{.014} \\
\midrule
\multicolumn{7}{@{}l}{\emph{Malignant subset, Dice}} \\
DbC (ours) & \vsd{.083^{*}}{.010} & \vsd{.113^{*}}{.018} & \vsd{.153^{*}}{.021} & \vsd{.201^{*}}{.007} & \vsd{.231^{*}}{.012} & \vsd{.261^{*}}{.010} \\
Disease-VL$^{\dagger}$ & \vsd{.059}{.005} & \vsd{.104}{.006} & \vsd{.141}{.005} & \vsd{.185}{.006} & \vsd{.205}{.013} & \vsd{.228}{.009} \\
\midrule
\multicolumn{7}{@{}l}{\emph{Malignant subset, AUROC}} \\
DbC (ours) & \vsd{.732^{*}}{.034} & \vsd{.799^{*}}{.032} & \vsd{.858^{*}}{.013} & \vsd{.879^{*}}{.014} & \vsd{.893^{*}}{.012} & \vsd{.913^{*}}{.007} \\
Disease-VL$^{\dagger}$ & \vsd{.688}{.006} & \vsd{.783}{.007} & \vsd{.842}{.009} & \vsd{.871}{.003} & \vsd{.878}{.004} & \vsd{.896}{.008} \\
\bottomrule
\end{tabular}

\end{table}

\flushfloats
\subsection{MCT-LTDiag for each of its four diseases}
\label{app:mct}

Tables~\ref{tab:mctzs} and~\ref{tab:mctft} report the MCT-LTDiag results for each of the four diseases, and their macro means are the rows of Table~\ref{tab:mct}.

\begin{table}[!htb]
\caption{MCT-LTDiag for each of its four diseases in the zero-shot regime, where each disease is scored with the LODO checkpoint that never saw it. The 95\% confidence interval is under each value. Bold marks the best value and $^{\dagger}$ a matched ablation.}
\label{tab:mctzs}
\centering
\tablefont
\setlength{\tabcolsep}{4pt}
\begin{tabular}{@{}lcccccccc@{}}
\toprule
Model & \multicolumn{2}{c}{HCC} & \multicolumn{2}{c}{ICC} & \multicolumn{2}{c}{Metastasis} & \multicolumn{2}{c}{Hemangioma} \\
\cmidrule(lr){2-3}\cmidrule(lr){4-5}\cmidrule(lr){6-7}\cmidrule(l){8-9}
\emph{Localization} & Dice & IoU & Dice & IoU & Dice & IoU & Dice & IoU \\
\midrule
SegVol & \vci{.046}{.031,.063} & \vci{.028}{.019,.041} & \vci{.041}{.033,.051} & \vci{.023}{.018,.029} & \vci{.027}{.022,.034} & \vci{.015}{.012,.019} & \vci{.061}{.049,.076} & \vci{.037}{.029,.046} \\
SAT & \vci{.092}{.062,.124} & \vci{.067}{.044,.093} & \vci{.052}{.034,.076} & \vci{.034}{.021,.052} & \vci{.108}{.086,.132} & \vci{.071}{.056,.088} & \vci{.125}{.099,.155} & \vci{.087}{.069,.110} \\
VoxTell & \vci{.064}{.043,.088} & \vci{.043}{.029,.060} & \vci{.044}{.029,.063} & \vci{.029}{.019,.042} & \vci{.068}{.054,.085} & \vci{.045}{.035,.058} & \vci{.080}{.063,.099} & \vci{.051}{.039,.066} \\
Disease-VL$^{\dagger}$ & \vci{.128}{.098,.163} & \vci{.090}{.067,.117} & \vcib{.163}{.132,.201} & \vcib{.121}{.097,.151} & \vci{.060}{.048,.077} & \vci{.041}{.032,.052} & \vci{.062}{.047,.080} & \vci{.041}{.031,.054} \\
\addlinespace[3pt]
DbC (ours) & \vcib{.225}{.180,.271} & \vcib{.176}{.142,.219} & \vci{.141}{.108,.181} & \vci{.108}{.082,.142} & \vcib{.127}{.103,.159} & \vcib{.097}{.078,.121} & \vcib{.198}{.161,.242} & \vcib{.154}{.125,.191} \\
\midrule
\emph{Detection} & AUROC & AP & AUROC & AP & AUROC & AP & AUROC & AP \\
\midrule
SegVol & \vci{.539}{.472,.605} & \vci{.205}{.159,.262} & \vci{.413}{.346,.480} & \vci{.141}{.109,.183} & \vci{.494}{.443,.546} & \vci{.437}{.382,.500} & \vci{.508}{.441,.573} & \vci{.198}{.155,.251} \\
SAT & \vci{.558}{.490,.624} & \vcib{.236}{.176,.311} & \vci{.499}{.429,.571} & \vci{.185}{.139,.250} & \vci{.443}{.394,.494} & \vci{.421}{.367,.488} & \vci{.612}{.548,.675} & \vcib{.281}{.213,.371} \\
VoxTell & \vci{.482}{.415,.549} & \vci{.211}{.162,.269} & \vci{.414}{.347,.482} & \vci{.151}{.114,.197} & \vci{.438}{.388,.489} & \vci{.405}{.352,.467} & \vci{.530}{.464,.595} & \vci{.185}{.143,.238} \\
Disease-VL$^{\dagger}$ & \vcib{.615}{.552,.678} & \vci{.201}{.170,.236} & \vci{.610}{.545,.671} & \vci{.225}{.170,.293} & \vci{.529}{.478,.579} & \vci{.420}{.369,.481} & \vci{.620}{.556,.676} & \vci{.265}{.205,.340} \\
\addlinespace[3pt]
DbC (ours) & \vci{.581}{.512,.648} & \vci{.188}{.149,.243} & \vcib{.666}{.593,.742} & \vcib{.364}{.269,.482} & \vcib{.676}{.626,.723} & \vcib{.490}{.415,.566} & \vcib{.635}{.576,.691} & \vci{.262}{.205,.337} \\
\bottomrule
\end{tabular}

\end{table}

\begin{table}[!htb]
\caption{MCT-LTDiag for each of its four diseases after full training on the internal data. The 95\% confidence interval is under each value. Bold marks the best value and $^{\dagger}$ a matched ablation.}
\label{tab:mctft}
\centering
\tablefont
\setlength{\tabcolsep}{4pt}
\begin{tabular}{@{}lcccccccc@{}}
\toprule
Model & \multicolumn{2}{c}{HCC} & \multicolumn{2}{c}{ICC} & \multicolumn{2}{c}{Metastasis} & \multicolumn{2}{c}{Hemangioma} \\
\cmidrule(lr){2-3}\cmidrule(lr){4-5}\cmidrule(lr){6-7}\cmidrule(l){8-9}
\emph{Localization} & Dice & IoU & Dice & IoU & Dice & IoU & Dice & IoU \\
\midrule
SegVol & \vci{.204}{.166,.245} & \vci{.151}{.122,.184} & \vci{.057}{.041,.078} & \vci{.038}{.027,.053} & \vci{.196}{.165,.232} & \vci{.153}{.128,.182} & \vci{.074}{.056,.096} & \vci{.050}{.037,.067} \\
SAT & \vci{.121}{.088,.155} & \vci{.083}{.060,.108} & \vci{.035}{.023,.050} & \vci{.023}{.014,.034} & \vci{.097}{.078,.120} & \vci{.065}{.051,.082} & \vci{.098}{.072,.127} & \vci{.072}{.053,.097} \\
VoxTell & \vci{.179}{.143,.218} & \vci{.127}{.101,.159} & \vci{.083}{.060,.113} & \vci{.058}{.040,.080} & \vci{.147}{.124,.176} & \vci{.104}{.088,.124} & \vci{.128}{.102,.158} & \vci{.089}{.071,.111} \\
Disease-VL$^{\dagger}$ & \vci{.182}{.148,.218} & \vci{.133}{.106,.164} & \vci{.160}{.123,.206} & \vci{.127}{.097,.166} & \vci{.177}{.148,.215} & \vci{.132}{.110,.160} & \vci{.214}{.176,.260} & \vci{.177}{.144,.219} \\
\addlinespace[3pt]
DbC (ours) & \vcib{.302}{.263,.346} & \vcib{.254}{.221,.292} & \vcib{.250}{.217,.293} & \vcib{.219}{.191,.256} & \vcib{.255}{.226,.293} & \vcib{.217}{.194,.246} & \vcib{.337}{.293,.390} & \vcib{.298}{.259,.346} \\
\midrule
\emph{Detection} & AUROC & AP & AUROC & AP & AUROC & AP & AUROC & AP \\
\midrule
SegVol & \vci{.646}{.578,.713} & \vci{.371}{.283,.471} & \vci{.548}{.479,.617} & \vci{.208}{.154,.278} & \vci{.554}{.503,.605} & \vci{.508}{.445,.576} & \vci{.706}{.641,.767} & \vci{.438}{.346,.544} \\
SAT & \vci{.626}{.562,.689} & \vci{.275}{.209,.361} & \vci{.534}{.466,.602} & \vci{.202}{.149,.279} & \vci{.447}{.395,.497} & \vci{.421}{.369,.485} & \vci{.779}{.718,.836} & \vci{.501}{.400,.613} \\
VoxTell & \vci{.623}{.551,.693} & \vci{.322}{.244,.424} & \vci{.570}{.503,.640} & \vci{.219}{.162,.291} & \vci{.464}{.412,.515} & \vci{.414}{.363,.473} & \vci{.749}{.687,.809} & \vci{.524}{.422,.623} \\
Disease-VL$^{\dagger}$ & \vci{.638}{.568,.706} & \vci{.338}{.249,.434} & \vci{.736}{.675,.794} & \vci{.339}{.260,.439} & \vci{.529}{.476,.582} & \vci{.480}{.417,.553} & \vcib{.963}{.933,.987} & \vci{.906}{.843,.957} \\
\addlinespace[3pt]
DbC (ours) & \vcib{.658}{.585,.731} & \vcib{.413}{.319,.522} & \vcib{.757}{.697,.817} & \vcib{.393}{.301,.499} & \vcib{.596}{.542,.648} & \vcib{.563}{.493,.636} & \vci{.962}{.923,.991} & \vcib{.913}{.829,.978} \\
\bottomrule
\end{tabular}

\end{table}

\flushfloats
\subsection{Localization and lesion size}
\label{app:size}

Small lesions are a known difficulty of lesion detection and localization in liver CT, where automated methods find and delineate lesions below 10~mm far less reliably than larger ones \citep{bilic2023lits, vorontsov2019liver}. Table~\ref{tab:size} measures this dependence for the unseen diseases of Section~\ref{sec:unseen}. The lesions of the six unseen diseases are pooled, each scored by the LODO checkpoint that never saw its disease. The analysis keeps the CTs with a single lesion of the unseen disease, for which the number of tokens of the lesion is known exactly. Size is counted in tokens of the grid on which every model is scored (Section~\ref{app:data}). Every model localizes small lesions far less accurately than large ones, and DbC is the most accurate in every size range.

The dependence on lesion size is in line with cyst being the hardest unseen disease to localize for every model (Section~\ref{sec:unseen}). Cysts are by far the smallest lesions of the six diseases, with a mean of 23 tokens per lesion against 152 to 543 for the other five, and most of them fall in the two smallest size ranges of Table~\ref{tab:size}.

\begin{table}[!htb]
\caption{Localization of unseen diseases by lesion size. Lesion Dice under leave-one-disease-out, pooled over the six diseases and grouped by the number of tokens that a lesion covers. Bold marks the best value and $^{\dagger}$ a matched ablation.}
\label{tab:size}
\centering
\footnotesize
\setlength{\tabcolsep}{5pt}
\begin{tabular}{@{}lcccccc@{}}
\toprule
Size (tokens) & Lesions & SegVol & SAT & VoxTell & Disease-VL$^{\dagger}$ & DbC (ours) \\
\midrule
4 or fewer & $78$ & $.002$ & $.000$ & $.000$ & $.013$ & $\mathbf{.018}$ \\
5 to 16 & $247$ & $.002$ & $.014$ & $.004$ & $.004$ & $\mathbf{.039}$ \\
17 to 64 & $245$ & $.193$ & $.165$ & $.057$ & $.095$ & $\mathbf{.213}$ \\
65 to 256 & $361$ & $.046$ & $.335$ & $.238$ & $.341$ & $\mathbf{.523}$ \\
More than 256 & $469$ & $.117$ & $.379$ & $.406$ & $.415$ & $\mathbf{.669}$ \\
\bottomrule
\end{tabular}

\end{table}

\flushfloats
\section{Further qualitative cases}
\label{app:cases}

\begin{figure}[!htb]
\centering
\includegraphics[width=0.85\textwidth]{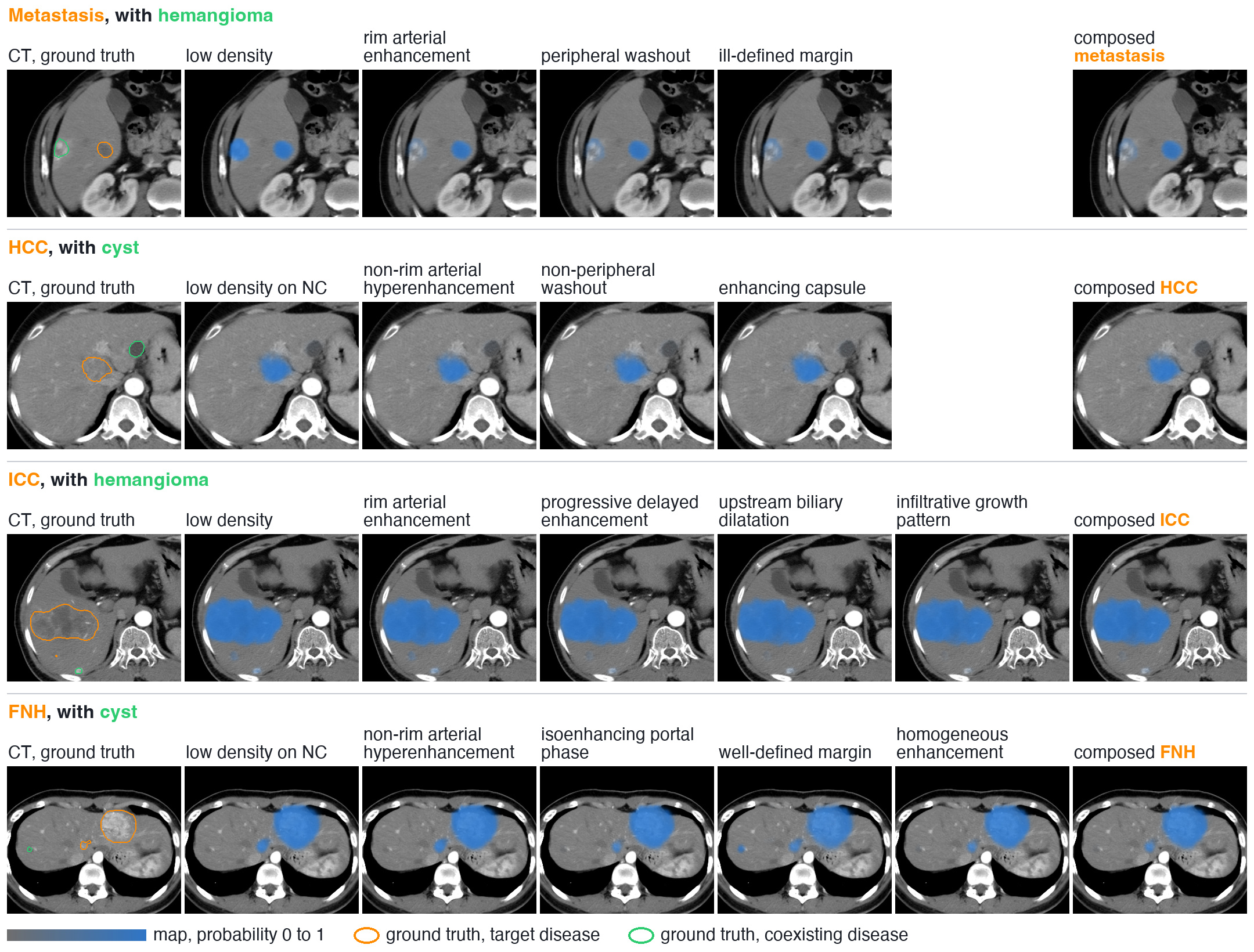}
\caption{Further qualitative cases of composition, each on one arterial-phase slice. A row shows the CT with the lesions of the target disease outlined in orange and the coexisting lesion in green, the maps of the required findings of the target disease and the composed diagnosis map. In the first case low density is active on both lesions, the other required findings of metastasis are active on the metastasis and only faintly on the hemangioma, and the diagnosis map follows them. In the second and third cases the required findings are active on the target lesion, and the diagnosis map stays low on the coexisting lesion. In the fourth case well-defined margin, which the compositions of FNH and cyst share, is also active on the cyst, whereas the diagnosis map is high on the FNH lesions only.}
\label{fig:cases}
\end{figure}

Figure~\ref{fig:cases} shows four further qualitative cases in the layout of Figure~\ref{fig:qualitative}, with metastasis, HCC, ICC and FNH as the target disease, so that the two figures together hold one case for each of the six diseases. Each CT holds a lesion of the target disease and a coexisting lesion of another disease on the same slice. As in Figure~\ref{fig:qualitative}, a finding that two compositions share can be active on both lesions, while the diagnosis map is high only where every required finding of the target disease is.

\flushfloats
\section{Code}
\label{app:code}

\begin{sloppypar}
Our code is available at \url{https://anonymous.4open.science/r/DiagnosisByComposition}. The code implements the DbC model of Section~\ref{sec:localizer} (\texttt{dbc/model.py}), the Dice-focal loss (\texttt{dbc/losses.py}), the stratified split and the sampling of training batches (\texttt{dbc/data.py}), the metrics (\texttt{dbc/metrics.py}), and the scripts that train a model (\texttt{train.py}) and evaluate a checkpoint (\texttt{evaluate.py}). The file \texttt{findings.json} holds the 40 findings with their contrast phases and the compositions of the lesion classes of Section~\ref{app:metrics}. The default settings are those of Table~\ref{tab:recipe}, and \texttt{README.md} describes the expected layout of the data and the two commands.
\end{sloppypar}

\end{document}